\documentclass[11pt]{article}

\usepackage[final]{acl}

\usepackage{times}
\usepackage{latexsym}

\usepackage[T1]{fontenc}

\usepackage[utf8]{inputenc}

\usepackage{microtype}

\usepackage{inconsolata}

\usepackage{graphicx}
\graphicspath{{figures/}}
\usepackage{subcaption}
\usepackage{pgfplots}
\pgfplotsset{compat=1.18}
\usepgfplotslibrary{groupplots}
\usetikzlibrary{arrows.meta}
\usepackage{xcolor}
\usepackage{pifont} %
\usepackage{iftex}
\ifXeTeX
  \usepackage{xeCJK}
\fi
\pgfplotsset{cnyaxis/.style={
    axis line style={gray!55},
    tick style={gray!55},
    xtick pos=bottom, ytick pos=left,
    grid style={dashed, gray!25},
    every axis plot/.append style={mark options={solid}},
}}
\usepackage{algorithm}
\usepackage{algpseudocode}
\usepackage{amsmath, amssymb, bm}
\usepackage{booktabs}
\usepackage{multirow}
\usepackage{listings}
\usepackage[most]{tcolorbox}
\definecolor{tagwalk}{HTML}{1A5FB4}    %
\definecolor{tagthink}{HTML}{8A8A8A}   %
\definecolor{taganswer}{HTML}{3F7A5C}  %
\definecolor{tagwrong}{HTML}{8D463F}   %
\definecolor{taginfo}{HTML}{B9770E}    %
\definecolor{promptbg}{HTML}{F7F9FC}   %
\definecolor{prompttitle}{HTML}{1A5FB4}%
\lstdefinestyle{cnyprompt}{
  basicstyle=\ttfamily\footnotesize,
  breaklines=true, breakautoindent=false, breakindent=0pt,
  columns=fullflexible, keepspaces=true, extendedchars=true,
  showstringspaces=false, aboveskip=0pt, belowskip=0pt,
  literate=
    {<thinking>}{{\color{tagthink}\itshape<thinking>}}{10}
    {</thinking>}{{\color{tagthink}\itshape</thinking>}}{11}
    {<walk>}{{\color{tagwalk}\bfseries<walk>}}{6}
    {</walk>}{{\color{tagwalk}\bfseries</walk>}}{7}
    {<answer>}{{\color{taganswer}\bfseries<answer>}}{8}
    {</answer>}{{\color{taganswer}\bfseries</answer>}}{9}
    {<information>}{{\color{taginfo}\bfseries<information>}}{13}
    {ü}{{\"u}}1 {é}{{\'e}}1 {ä}{{\"a}}1 {ö}{{\"o}}1 {ß}{{\ss}}1,
}

\newcommand{\tval}{\textcolor{gray!60}{\textendash}}

\newsavebox{\cnyfitbox}
\newcommand{\fitwidth}[2]{%
  \sbox{\cnyfitbox}{#2}%
  \ifdim\wd\cnyfitbox>#1\relax
    \resizebox{#1}{!}{\usebox{\cnyfitbox}}%
  \else
    \usebox{\cnyfitbox}%
  \fi}

\usepackage{etoolbox}
\AtBeginEnvironment{table}{\nolinenumbers}
\AtBeginEnvironment{table*}{\nolinenumbers}
\AtBeginEnvironment{figure}{\nolinenumbers}
\AtBeginEnvironment{figure*}{\nolinenumbers}
\AtBeginEnvironment{algorithm}{\nolinenumbers}

\title{Call Neighbours Yourself: Graph Walks with Destination-Conditioned On-Policy Self-Distillation}

\author{Yilun Liu, Boyu Luo, Yanran Tang, Ruihong Qiu and Zi Huang \\
  School of Electrical Engineering and Computer Science \\
  The University of Queensland \\
  Brisbane, Queensland, Australia \\
  \texttt{\{yilun.liu, boyu.luo, yanran.tang, r.qiu, helen.huang\}@uq.edu.au}}

\begin{document}
\maketitle
\begin{abstract}
Reasoning over text-attributed graphs (TAGs) requires large language models (LLMs) to combine a node's text with evidence distributed across its neighbourhood. Existing methods fix the set of accessible neighbours before generation, forcing reasoning to operate over a static context and preventing the model from acquiring missing evidence during inference. We argue that neighbour selection should itself be part of the reasoning process. To this end, we propose \textbf{C}all \textbf{N}eighbours \textbf{Y}ourself (CNY), a framework that enables LLMs to proactively explore graph neighbourhoods through topology-constrained graph-walk actions. Instead of reasoning over a pre-selected neighbour set, CNY exposes lightweight neighbour previews and learns when to expand candidate neighbours for additional evidence. To address the delayed-credit challenge of neighbour exploration, we introduce destination-conditioned on-policy self-distillation, which retrospectively evaluates a selected neighbour after its content is revealed and converts the resulting change in action preference into an action-level training signal. Experiments on standard TAG reasoning benchmarks under a unified raw-text setting show that CNY consistently outperforms fixed-context post-training baselines. Furthermore, the learned exploration policy transfers to unseen graphs and to a graph-level task not encountered during training. Code is available at \url{https://github.com/superallen13/CNY}.
\end{abstract}

\section{Introduction}

\begin{figure}[!t]
\centering
\input{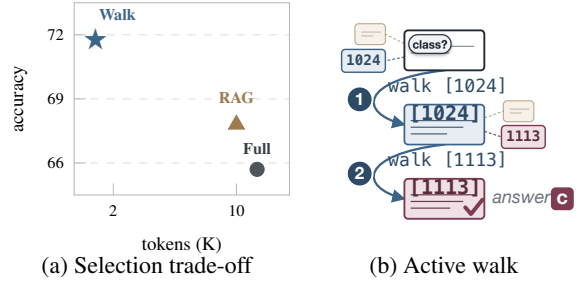}
\caption{\textbf{CNY at a glance.} (a) On WikiCS targets whose full neighbourhood fits in a $32$K context, a frozen Qwen2.5-32B that selects neighbours during generation (\textbf{Walk}) outperforms a retrieval-augmented generation pipeline (\textbf{RAG}) that embeds the target's text, retrieves its top-$10$ neighbours and concatenates them as context, at $6.3\times$ fewer tokens. (b) CNY learns this selection as a graph-walk action, replacing the fixed pre-selected neighbour context of prior work.}
\label{fig:motivation}
\end{figure}

Citation networks, hyperlink graphs, and product co-purchase networks are text-attributed graphs (TAGs), where each node contains free-form text and is connected to other text-bearing nodes~\cite{wu2025llmnodebed,casegnn,caselink,lexa,tntood,cassette}. Reasoning over TAGs therefore requires a model to use not only the target node's own text, but also evidence from its graph neighbourhood~\cite{he2024tape,li2024zerog}. Existing large language models (LLMs)-based TAG methods usually turn this requirement into a prompt construction problem: a subset of neighbouring texts is selected before generation and then provided as fixed context~\cite{tang2024graphgpt,chen2024llaga,kong2025gofa}, which is adopted by recent post-training methods, such as Graph-R1~\cite{wu2025graphr1} and TRN-R1-Zero~\cite{liu2026trnr1zero}. Although they optimise how the LLM reasons over the provided context, neighbour acquisition itself remains decoupled from reasoning: the LLM cannot adapt its evidence gathering to what it discovers as it reasons.

This static-context assumption becomes problematic on dense TAGs. Neighbourhoods often exceed practical context limits: on WikiCS the full one-hop neighbourhood already exceeds a $32$K-token window for $40.3\%$ of the $2{,}340$ test targets, forcing systems to read only a subset. Yet which neighbours are selected largely determines prediction quality, since relevant evidence is easily buried among irrelevant nodes or overlooked when placed mid-context~\cite{liu2024lostmiddle,hsieh2024ruler,bai2024longbench}. Once a subset is fixed before generation, the LLM cannot recover if decisive evidence lies elsewhere.

This paper argues that neighbour selection should be part of the reasoning process itself. Instead of passively consuming a fixed neighbour set, an LLM should first observe lightweight previews of candidate neighbours, such as a title, a short summary, or a fixed-length text prefix, and then decide which neighbours deserve deeper inspection. This interleaves reasoning and neighbour acquisition rather than separating them as in retrieval-then-generate pipelines. As in Figure~\ref{fig:motivation}, our motivation study supports this view: on WikiCS targets whose full neighbourhoods fit in context (so neither method is forced to truncate), a frozen Qwen2.5-32B-Instruct model that selects neighbours during generation outperforms a RAG pipeline that selects neighbours by embedding similarity, using the same backbone and identical neighbour previews, while using substantially fewer tokens. However, training an LLM to walk well is non-trivial: the usefulness of a walk can only be judged after its destination is revealed and folded into subsequent reasoning, so an outcome-only RL objective has no direct signal to credit individual walks.

This paper proposes \textbf{C}all \textbf{N}eighbours \textbf{Y}ourself (CNY), which trains the LLM to issue a \texttt{<walk>} action mid-generation to read a specific neighbour, replacing the pre-selected neighbour set with evidence the LLM acquires step by step. Process reward models~\cite{lightman2024prm800k,wang2024mathshepherd} would normally supply the missing per-walk credit, but they require step labels that graph walks do not provide. CNY therefore introduces destination-conditioned \textbf{O}n-\textbf{P}olicy \textbf{S}elf-\textbf{D}istillation (OPSD): once a walk reveals its destination, the same base LLM re-scores its own walk action, and the resulting probability shift is used as per-action credit, with no step labels, no external judge, and no extra rollouts. Our contributions are summarised as follows:
\begin{itemize}
    \item We identify static neighbour selection as a key limitation of LLM-based reasoning on dense TAGs and formulate neighbour inspection as an adaptive evidence-acquisition problem.
    \item We propose CNY, a proactive neighbour exploration framework that enables LLMs to select and expand neighbours during generation through graph-walk actions.
    \item We introduce a destination-conditioned on-policy self-distillation strategy that assigns action-level credit to neighbour-selection decisions without step-level annotations.
    \item Extensive experiments across citation, hyperlink, co-purchase and knowledge-graph benchmarks demonstrate that CNY surpasses fixed-context baselines.
\end{itemize}

\section{Related Work}
\label{sec:related}
Related work spans three threads. Existing \textbf{LLM-based reasoning on text-attributed graphs} (TAGs) mainly encodes node text and graph structure through graph-aware prompting, adapters, or reinforcement learning over fixed neighbourhood contexts selected by non-learned heuristics~\cite{li2024zerog,he2024tape,tang2024graphgpt,chen2024llaga,kong2025gofa,chen2024graphwiz,wang2025npgmuse,wu2025graphr1,liu2026trnr1zero}. In parallel, \textbf{agentic retrieval systems} interleave reasoning with external search by issuing free-form queries to semantic retrievers over unstructured document corpora~\cite{asai2024selfrag,li2025searcho1,jin2025searchr1,song2025r1searcher,chen2025research}. In contrast, CNY constrains its actions to the graph topology rather than an open query space, and a walk observes the destination node's own text rather than a retrieval approximation. Finally, \textbf{sparse-reward reasoning optimisation} has been studied through process reward models requiring step-level supervision~\cite{lightman2024prm800k,wang2024mathshepherd} and on-policy self-distillation methods using outcome-privileged teachers~\cite{shenfeld2026sdft,hubotter2026sdpo,zhao2026opsd,wang2026openclaw}. In contrast, CNY formulates neighbour selection itself as a learnable graph walk and introduces destination-conditioned OPSD to provide action-level supervision from revealed neighbour destinations without labelled intermediate trajectories. Full discussion is provided in Appendix~\ref{app:related}.

\begin{figure*}[t]
\centering
\input{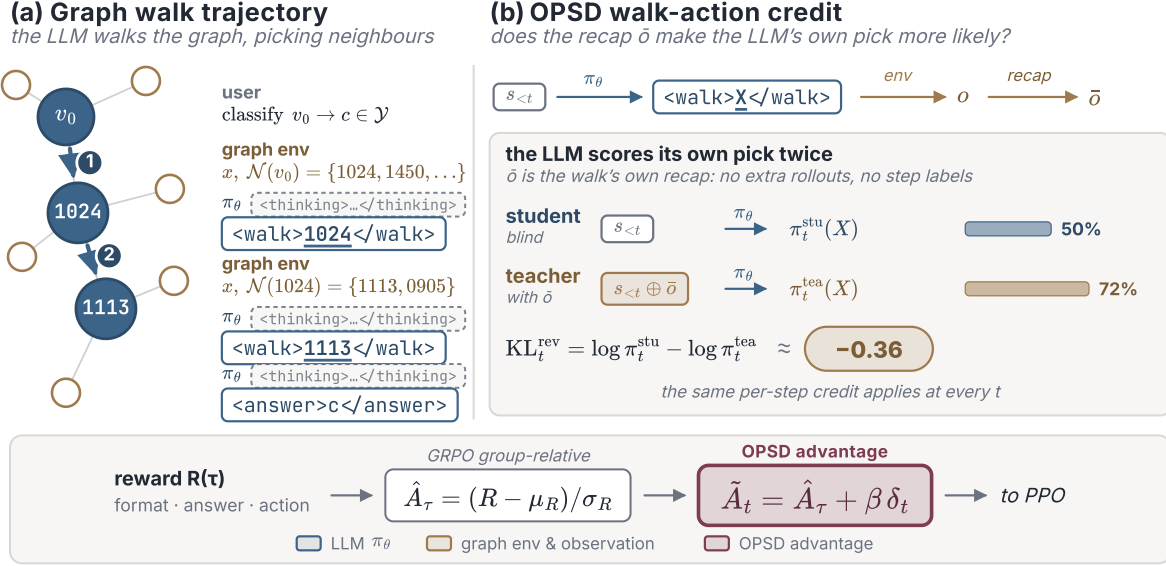}
\caption{\textbf{CNY graph walk trajectories and OPSD credit.} \textbf{(A)} The LLM emits a \texttt{<thinking>} span and a \texttt{<walk>}$X$\texttt{</walk>} action; the environment returns destination text and expands the frontier. The credit applies only to the node-ID tokens of successful walks. \textbf{(B)} OPSD replays the walk under $\pi_\theta$ as a destination-blind student and a recap-conditioned teacher; their per-token reverse KL shifts the GRPO advantage at walk tokens (red box).}
\label{fig:method}
\end{figure*}

\section{Problem Definition}
\label{sec:problem}

A text-attributed graph (TAG) is a tuple $\mathcal{G} = \{\bm{A}, \bm{X}, \mathcal{Y}\}$, where $\bm{A} \in \{0,1\}^{|\mathcal{V}| \times |\mathcal{V}|}$ is the adjacency matrix over node set $\mathcal{V}$, $\bm{X} = \{x_v\}_{v \in \mathcal{V}}$ assigns each node a textual description $x_v$, and $\mathcal{Y}$ is a finite label set with class descriptions $\{\mathrm{desc}(y)\}_{y \in \mathcal{Y}}$. Given a target node $v_0 \in \mathcal{V}$, the task is to predict its label $\hat y \in \mathcal{Y}$ from evidence distributed across its graph neighbourhood.

The model cannot assume access to the entire neighbourhood, so it must make a prediction under a limited reading budget where only a subset of the neighbourhood evidence is observed. Selecting that subset well governs the prediction: reading uninformative neighbours, or burying the decisive one among many, degrades the answer, and on dense graphs such as WikiCS the full one-hop text exceeds a 32K-token window for 40.3\% of evaluation nodes. The challenge is thus to identify and utilise the most informative evidence for prediction.

\section{Method: Call Neighbours Yourself}
\label{sec:method}

CNY formulates the neighbourhood evidence acquisition procedure as a sequential decision process trained with reinforcement learning (Figure~\ref{fig:method}). The LLM actively acquires evidence by walking the graph rather than simply reasoning over a fixed neighbour set (\S\ref{sec:rollout}); a trajectory-level reward scores both prediction correctness and interaction validity (\S\ref{sec:reward}); the model is optimised against that reward (\S\ref{sec:grpo}); and destination-conditioned on-policy self-distillation (OPSD) retrospectively evaluates each walk decision after observing the destination it retrieved, converting the resulting hindsight signal into token-level credit (\S\ref{sec:opsd}).

\subsection{Interactive Graph Environment}
\label{sec:rollout}

CNY treats graph reasoning as an interactive environment in which evidence is acquired on demand rather than provided as a fixed neighbourhood context. Initially, the LLM observes the ego-node text $x_{v_0}$, the label descriptions, and a short preview for each neighbour in the initial frontier. A neighbour's full text remains hidden until it is explicitly selected.

At reasoning step $k$, generation continues from the current context,
\begin{equation}
s_k=(q_0,\,h_{k-1},\,\mathcal{F}_{k-1}),
\end{equation}
where $q_0$ is the initial prompt, $h_{k-1}$ records previously revealed node contents and generated reasoning traces, and $\mathcal{F}_{k-1}$ is the frontier containing nodes currently available for inspection.

Executing \texttt{<walk>} on node $X$ reveals its full text $x_X$ and expands the frontier with previously unseen neighbours:
\begin{equation}
\mathcal{F}_k=
(\mathcal{F}_{k-1}\setminus\{X\})
\cup
(\mathcal{N}(X)\setminus\mathcal{V}^{\mathrm{read}}),
\label{eq:frontier-update}
\end{equation}
where $\mathcal{V}^{\mathrm{read}}$ denotes nodes whose full text has already been revealed. Consequently, every newly available node must be adjacent to a previously inspected node, forcing evidence acquisition to follow graph connectivity rather than access arbitrary nodes.

\subsection{Reward}
\label{sec:reward}

Sparse correctness rewards are ineffective for graph-walk learning because most trajectories are initially incorrect. Under GRPO, a group containing only incorrect trajectories yields zero advantage for every sample and therefore no learning signal (\S\ref{sec:grpo}). To provide learning signal before correct reasoning emerges, trajectories that satisfy desirable intermediate properties receive non-zero rewards.

The reward prioritises (i) correct prediction $c$, (ii) valid action formatting $f$, (iii) successful evidence acquisition $s$ (at least one frontier node is revealed), and (iv) answer completion $e$ (a valid \texttt{<answer>} is emitted):
\begin{equation}
R(\tau) =
\begin{cases}
1.0 & c{=}1,\ f{=}1, \\
0.6 & c{=}1,\ f{=}0, \\
0.3 & c{=}0,\ f{=}1,\ s{=}1, \\
0.2 & c{=}0,\ f{=}1,\ s{=}0, \\
0.1 & c{=}0,\ f{=}0,\ e{=}1, \\
0.0 & \text{otherwise.}
\end{cases}
\label{eq:reward-tier}
\end{equation}

The values are chosen only to realise the ordering
$1.0 > 0.6 > 0.3 > 0.2 > 0.1 > 0$,
corresponding to a lexicographic preference over desirable behaviours rather than calibrated utility estimates. (ablated in \S\ref{sec:opd-ablation}, Table~\ref{tab:reward-ablation}) The reward remains unchanged in the $\beta{=}0$ ablation (\S\ref{sec:opd-ablation}), ensuring that gains attributed to OPSD are not explained by reward shaping.

\subsection{Reinforcement Learning Optimisation}
\label{sec:grpo}

Each graph walk induces a trajectory $\tau$, and the policy $\pi_\theta$ is trained to maximise the expected trajectory reward
\begin{equation}
\max_\theta \;
\mathbb{E}_{\tau\sim\pi_\theta}
\bigl[R(\tau)\bigr].
\end{equation}

CNY adopts GRPO~\cite{shao2024deepseekmath} and its Dr.~GRPO implementation~\cite{drgrpo}, using asymmetric clipping and no KL-to-reference penalty. For each target node $v_0$, a group of $N$ trajectories $\{\tau_i\}_{i=1}^{N}$ is sampled and scored by Eq.~\ref{eq:reward-tier}. Rewards are standardised within the group to obtain the trajectory-level advantage
\begin{equation}
\hat A_{\tau_i}
=
\frac{R(\tau_i)-\mu_R}
{\sigma_R},
\label{eq:grpo-adv}
\end{equation}
where $\mu_R$ and $\sigma_R$ denote the group mean and standard deviation.

Vanilla GRPO broadcasts the same advantage $\hat A_{\tau_i}$ to every token in trajectory $\tau_i$ and optimises the resulting PPO-style objective~\cite{schulman2017ppo}. As discussed next, this uniform credit assignment is poorly matched to graph-walk trajectories because only a small subset of tokens determines which neighbours are explored. OPSD addresses this mismatch by re-pricing the trajectory advantage at walk decisions.

\subsection{Destination-Conditioned On-Policy Self-Distillation (OPSD)}
\label{sec:opsd}

GRPO assigns the same trajectory advantage to all tokens, yet the quality of a walk decision becomes observable only after the destination node is revealed. OPSD converts this hindsight information into a local credit signal for neighbour selection while remaining fully on-policy.

After a walk on node $X$ reveals observation $o$, we construct a bounded-length destination recap

\begin{equation}
\bar{o}=\varphi(o),
\qquad
|\bar{o}| \le L ,
\label{eq:recap}
\end{equation}

where $\varphi(\cdot)$ compresses the revealed destination content into a topic-level textual description while excluding node identifiers, class labels, and explicit walk recommendations. The recap serves as a compact representation of what was discovered at the destination and is generated by the same policy $\pi_\theta$ (Fig.~\ref{fig:recap-prompt}).

The realised walk action is then evaluated twice by $\pi_\theta$: once using the original context (\textbf{student}) and once with \texttt{[Destination: $\bar{o}$]} inserted immediately before the walk span (\textbf{teacher}). For a node-ID token $y_t$ in \texttt{<walk>}X\texttt{</walk>}, the resulting hindsight credit is

\begin{equation}
\delta_t
\triangleq
\log \pi^{\mathrm{tea}}_t
-
\log \pi^{\mathrm{stu}}_t ,
\label{eq:opsd-credit}
\end{equation}

where $\pi^{\mathrm{stu}}_t$ and $\pi^{\mathrm{tea}}_t$ denote the probabilities assigned to the realised token under the two conditioning contexts. A positive $\delta_t$ indicates that, after observing what was actually found at the destination, the model becomes more confident in the neighbour it previously selected.

OPSD augments the GRPO advantage only on walk-destination tokens:

\begin{equation}
\tilde A_t=
\begin{cases}
\hat A_\tau+\beta\delta_t,
&
t \in \text{walk action tokens},
\\
\hat A_\tau,
&
\text{otherwise},
\end{cases}
\label{eq:opsd-adv}
\end{equation}

where $\delta_t$ is detached from $\theta$ and $\beta$ controls the strength of the hindsight signal. Destination information provides an additive per-action credit on top of the trajectory-level advantage, applied only to the node-ID tokens of a successful walk; the outcome reward remains the sole supervision for all other tokens.

Reasoning tokens therefore remain supervised solely by the trajectory-level reward, avoiding the degradation of reasoning traces reported for self-distillation methods~\cite{kim2026whyselfdistill}. App.~\ref{app:revkl-deriv} shows that $\delta_t$ is the gradient of a per-token reverse-KL self-distillation objective~\cite{hubotter2026sdpo,zhao2026opsd}, making Eq.~\eqref{eq:opsd-adv} the gradient of the corresponding combined objective. Unlike prior self-distillation methods, the teacher conditions on the destination revealed by the realised walk itself rather than on a gold answer or external supervision.

\subsection{Training}
\label{sec:training}

For a target node $v_0$, each training step samples a group of $N$ rollouts from the current policy (\S\ref{sec:rollout}) and scores them with the trajectory reward $R(\tau)$ (\S\ref{sec:reward}). GRPO converts the group rewards into trajectory-level advantages $\hat A_\tau$ (\S\ref{sec:grpo}), which OPSD further refines on walk-destination tokens to obtain $\tilde A_t$ (\S\ref{sec:opsd}).

The policy is optimised with the PPO-style clipped objective
\begin{equation}
\mathcal{J}(\theta)
=
\mathbb{E}_{\tau}
\Bigl[
\frac{1}{|\tau|}
\sum_t
\min
\bigl(
\rho_t \tilde A_t,
\bar\rho_t \tilde A_t
\bigr)
\Bigr],
\label{eq:training-loss}
\end{equation}
where
\(
\rho_t
=
\pi_\theta(y_t|s_{<t})
/
\pi_{\theta_{\mathrm{old}}}(y_t|s_{<t})
\)
is the importance ratio and
\(
\bar\rho_t
=
\operatorname{clip}
(
\rho_t,
1-\epsilon_{\mathrm{lo}},
1+\epsilon_{\mathrm{hi}}
)
\)
applies the asymmetric clipping of Dr.~GRPO (\S\ref{sec:grpo}).

Training uses the multi-task mixture described in \S\ref{sec:setup}. The complete procedure is summarised in Algorithm~\ref{alg:cny-train} (App.~\ref{app:algorithms}).

\begin{table*}[t!]
\centering
\footnotesize
\setlength{\tabcolsep}{4pt}
\resizebox{\textwidth}{!}{%
\begin{tabular}{l | cc cc cc | c | c}
\toprule
 & \multicolumn{6}{c|}{Node-level} & Edge-level & Graph-level \\
\cmidrule(lr){2-7} \cmidrule(lr){8-8} \cmidrule(lr){9-9}
Task & \multicolumn{2}{c}{Cora} & \multicolumn{2}{c}{WikiCS} & \multicolumn{2}{c|}{Products} & FB15K237 & \multirow{2}{*}{Expla-Graph} \\
\cmidrule(lr){2-3} \cmidrule(lr){4-5} \cmidrule(lr){6-7} \cmidrule(lr){8-8}
 & 7-way & 2-way & 10-way & 5-way & 10-way & 5-way & 10-way & \\
\midrule
Llama2-7B   & 47.92 & 73.45 & 40.10 & 58.77 & 58.71 & 64.33 & 48.32 & 57.76 \\
Mistral-7B  & 60.54 & 88.39 & 63.63 & 71.90 & 70.16 & 74.94 & 62.48 & 68.77 \\
\midrule
OFA (7B)        & 28.65 & 56.92 & 21.20 & 35.15 & 30.43 & 39.31 & --    & 51.36 \\
UniGraph (7B)   & 69.53 & \underline{89.74} & 43.45 & 60.23 & 66.07 & 75.73 & --    & --    \\
LLaGA (7B)      & 51.85 & 62.73 & --    & --    & 34.15 & 39.72 & --    & --    \\
GOFA-T (7B)     & 70.81 & 85.73 & 71.17 & 80.93 & 79.33 & 87.13 & 73.59 & 79.49 \\
GOFA-F (7B)     & 69.41 & 87.52 & 68.84 & 80.52 & 80.03 & 88.34 & \underline{80.69} & 71.34 \\
TRN-R1-Zero (7B)  & 72.59 & 85.93 & 73.63 & 78.25 & 81.10 & 88.00 & 74.70 & 85.92 \\
Graph-R1 (14B)  & 68.15 & 89.08 & 73.25 & 79.62 & 85.72 & \underline{91.78} & 75.17 & \underline{89.71} \\
\midrule
\textsc{CNY} (7B) & \textbf{75.37} & 89.63 & \underline{74.32} & \underline{81.20} & \underline{86.00} & 90.30 & 76.41 & 88.45 \\
\textsc{CNY} (14B) & \underline{73.70} & \textbf{91.48} & \textbf{76.75} & \textbf{85.51} & \textbf{87.30} & \textbf{91.87} & \textbf{82.13} & \textbf{92.60} \\
\bottomrule
\end{tabular}%
}

\caption{\textbf{Zero-shot accuracy (\%, $\uparrow$).} LLM and graph-foundation rows are from Graph-R1~\cite{wu2025graphr1}; Graph-R1 is re-evaluated under the protocol of \S\ref{sec:setup}. \textbf{Bold}: best; \underline{underline}: second-best. CNY's lead over the runner-up reasoner is significant at $p\le0.05$ (paired $t$-test on per-instance correctness, Bonferroni-corrected).}
\label{tab:main}
\end{table*}

\section{Experiments}
\label{sec:experiments}

\subsection{Setup}
\label{sec:setup}

\paragraph{Datasets.} CNY is trained on a multi-task mixture of eight text-attributed graphs spanning citation, e-commerce, hyperlink and knowledge-graph domains: seven for node classification~\cite{cat,puma,gcondenser,host,gfmate,tide} and one (WN18RR) for relation classification. Following the zero-shot protocol of Graph-R1~\cite{wu2025graphr1}, a single checkpoint is then evaluated without per-task fine-tuning on five held-out graphs spanning node-, edge- and graph-level tasks, including the citation graphs Cora~\cite{sen2008cora} and WikiCS~\cite{mernyei2020wikics} and the co-purchase graph ogbn-products~\cite{hu2020ogb}; graph-level reasoning is held out from training, so the graph-level evaluation (Expla-Graph) probes cross-task transfer, as does the open-ended KGQA evaluation of \S\ref{sec:webqsp}. The suite is inherited unchanged from the GOFA-aligned benchmark of Graph-R1 and TRN-R1-Zero~\cite{kong2025gofa,liu2026trnr1zero}; per-dataset statistics are in App.~\ref{app:datasets}. All scores are accuracy under greedy decoding (temperature $0$). Each neighbour preview is a deterministic prefix of the node's raw text truncated to a per-dataset token budget, with no summarisation model or metadata involved (App.~\ref{app:preview-spec}).

\paragraph{Baselines.} Baselines span three families: general-purpose LLMs, graph foundation models, and RL-post-trained reasoners (Graph-R1 and TRN-R1-Zero~\cite{liu2026trnr1zero}), listed in full in App.~\ref{app:setup-details}. The two reasoners and \textsc{CNY} are evaluated on identical raw node text, so the gap reflects the learned walk rather than the inputs.

\paragraph{Models.} Main results use Qwen2.5-14B-Instruct~\cite{qwen25} as the CNY backbone, matched in scale to the strongest reasoning baseline (Graph-R1, a DeepSeek-R1~\cite{guo2025deepseekr1}-distilled Qwen2.5-14B) so the lead cannot be attributed to a larger backbone; TRN-R1-Zero is evaluated at its released 7B scale. Hyperparameters and hardware are listed in App.~\ref{app:implementation}. CNY uses OPSD with $\beta{=}0.03$ held fixed across all datasets; the $\beta{=}0$ GRPO ablation (same reward, Eq.~\ref{eq:reward-tier}) is studied in \S\ref{sec:opd-ablation}. The base-model study (\S\ref{sec:capability-gap}) trains CNY from five backbones spanning two families and four scales (Llama-3.2-3B~\cite{llamateam2024llama3} and Qwen 4B--14B).

\subsection{Main Results}
\label{sec:main-results}

Held-out zero-shot accuracy is reported in Table~\ref{tab:main}, evaluating CNY on two transfer axes: to unseen graphs of a trained task family, and to a task family never seen during training. \textbf{CNY attains the highest accuracy on every setting in Table~\ref{tab:main}, leading the strongest prior reasoner, Graph-R1, on every task family.} The prior reasoners answer an oversized neighbourhood by fixing the context before generation, through summarisation or random sampling, whereas CNY walks to the neighbours it needs during reasoning. The two axes are examined in turn.

\paragraph{Cross-dataset transfer (seen task types).} On the four datasets whose task family is represented in training, CNY leads every graph foundation model and reasoning baseline, and the size of the lead tracks how much the prediction depends on selecting the right neighbours. On the dense WikiCS graph CNY reaches $76.8$ at $10$ ways and $85.5$ at $5$ ways, ahead of the second-best $73.6$ and $80.9$, whereas on the sparse Cora graph, whose full neighbourhood already fits in the prompt, the margin narrows to $73.7$ against $72.6$. On Products CNY reaches $87.3$ against $85.7$, and on the relation-classification graph FB15K237, where a single decisive neighbour settles the label, $82.1$ against $80.7$. This ordering matches the premise of \S\ref{sec:method}: a learned reading LLM helps most when the neighbourhood is large and selecting the right evidence matters most.

\paragraph{Cross-task transfer (unseen task types).} Expla-Graph is the graph-level cross-task evaluation, a binary stance judgement over a per-instance commonsense explanation graph, a task family absent from training. Its concepts form a walkable neighbourhood, so the same multi-step walk template applies: CNY is shown only the seed concepts and must walk to reveal the rest, whereas the reasoning baselines (Graph-R1, TRN-R1-Zero) receive the full explanation graph in context. Despite this information handicap, and without any graph-level reasoning during training, CNY attains the strongest stance accuracy ($92.60$ against $89.71$ for Graph-R1 and $85.92$ for TRN-R1-Zero, Table~\ref{tab:main}). \textbf{The learned walk LLM therefore transfers to a graph-level task it never saw during training.}

\subsection{Effectiveness of Walking}
\label{sec:walk-effectiveness}

Does the gain come from adaptive neighbour selection, or simply from putting more text in context? We compare the frozen CNY-14B model against itself with the walk disabled, under three controls on the full held-out splits.

\paragraph{Matched text budget.} Each node is answered either directly from one 1-hop neighbour's full text, or by walking to selected neighbours at the same text budget. Walking raises accuracy on every dataset (Table~\ref{tab:walk-vs-direct}), so the gain comes from reading selectively rather than from more content in context.

\paragraph{Matched inference compute.} Under a comparable inference token budget on WikiCS, one greedy walk ($77.0$ at $2{,}341$ tokens per node) outperforms both self-consistency over four samples ($74.5$) and best-of-$4$ reranking ($74.4$, each at $2{,}899$ tokens). Additional test-time compute therefore does not explain the walk's margin, nor does preview exposure, since showing every neighbour's preview while forbidding the walk (preview+direct) improves the direct baseline only modestly (App.~\ref{app:preview-exposure}).

\paragraph{Degree-preserving rewire.} If the model were doing only semantic retrieval over neighbour text, randomising the graph while keeping every node's degree fixed should leave accuracy unchanged. On WikiCS we apply random edge swaps, preserving degrees but destroying topology, with texts and labels untouched. Walk accuracy collapses on WikiCS ($76.6\!\to\!73.0$, matching the preview+direct baseline at $73.4$) while walks-per-node rises ($1.24\!\to\!2.01$), consistent with recognising that each step now returns less useful evidence. The walk also lands on same-class neighbours well above the homophily rate ($63.4\%\!\to\!74.0\%$ on WikiCS, $76.6\%\!\to\!84.0\%$ on Cora), yet rarely on the most embedding-similar preview ($22\%$ of picks, mean similarity rank $4.0$ against $1.0$ for a retriever). \textbf{Selective reading over real graph structure, not preview exposure or the act of walking itself, is what makes the walk LLM useful.}

\begin{table}[t]
\centering
\fitwidth{\columnwidth}{\definecolor{cnyHero}{HTML}{7D3C52}
\definecolor{cnyDrop}{HTML}{888888}
\setlength{\tabcolsep}{3pt}
\begin{tabular}{llrrrr}
\toprule
Dataset & Walks & Nodes & Direct & Walk & $\Delta$ \\
\midrule
\multirow{2}{*}{WikiCS}  & 1 walk & 78\% & 75.7 & \textbf{81.6} & \textcolor{cnyHero}{$+5.9$} \\
 & $\geq$2 walks & 22\% & 54.2 & \textbf{58.7} & \textcolor{cnyHero}{$+4.5$} \\
\midrule
\multirow{2}{*}{Cora} & 1 walk & 70\% & 75.0 & \textbf{82.4} & \textcolor{cnyHero}{$+7.4$} \\
 & $\geq$2 walks & 30\% & 42.1 & \textbf{54.9} & \textcolor{cnyHero}{$+12.8$} \\
\midrule
\multirow{2}{*}{Products} & 1 walk & 92\% & 86.5 & \textbf{91.1} & \textcolor{cnyHero}{$+4.7$} \\
 & $\geq$2 walks & 8\% & 42.3 & \textbf{48.0} & \textcolor{cnyHero}{$+5.7$} \\
\bottomrule
\end{tabular}
}
\caption{\textbf{Walking improves accuracy across datasets.} Frozen CNY-14B at a matched text budget ($1.1$--$1.3$ neighbours per node). \textbf{Direct}: one 1-hop neighbour's full text; \textbf{Walk}: walk to selected neighbours; rows split by walk count, \textbf{Nodes} is the bucket share. Setup and controls in \S\ref{sec:walk-effectiveness}.}
\label{tab:walk-vs-direct}
\end{table}

\subsection{Generalisation to Multi-Hop KGQA}
\label{sec:webqsp}

Realised walk depth saturates near $1.2$ walks per node on the classification benchmarks, leaving open whether the policy generalises to tasks demanding deeper exploration. On the multi-hop KGQA benchmark WebQSP~\cite{yih2016webqsp} (all $1{,}628$ test questions, substring Hits@$1$ protocol of G-Retriever~\cite{he2024gretriever}), the frozen CNY-14B walks each question's Freebase subgraph with no question-answering training. Walking reaches $58.4$ against $38.5$ for answering from the identical previews ($+19.9$), and neither the question alone ($44.0$, parametric memory) nor the full 1-hop neighbourhood concatenated in context ($49.0$) closes the gap (Table~\ref{tab:webqsp}). The policy also walks deeper unprompted, at $2.31$ walks per question. \textbf{The learned walk policy scales its exploration to the demands of the task and transfers zero-shot to multi-hop question answering.}

\begin{table}[t]
\centering
\footnotesize
\setlength{\tabcolsep}{4pt}
\begin{tabular}{lc}
\toprule
Settings & Hits@$1$ \\
\midrule
CNY-14B, walk ($T_{\max}{=}5$)  & $\mathbf{58.4}$ \\
CNY-14B, full 1-hop context & $49.0$ \\
CNY-14B, question only (no graph) & $44.0$ \\
CNY-14B, direct from previews   & $38.5$ \\
\bottomrule
\end{tabular}
\caption{\textbf{Zero-shot multi-hop KGQA on WebQSP.} One frozen checkpoint, no question-answering training; all variants share subgraphs, prompts and scoring.}
\label{tab:webqsp}
\end{table}

\subsection{Effectiveness of OPSD}
\label{sec:opd-ablation}

This experiment isolates the contribution of the OPSD credit from outcome-only reward. The same backbone (CNY-7B) is trained twice on the multi-task mixture with identical reward (Eq.~\ref{eq:reward-tier}) and rollouts, differing only in whether the OPSD term is on ($\beta{=}0.03$) or off ($\beta{=}0$, the GRPO ablation), and held-out accuracy is tracked against training reward throughout. The $\beta$-sweep below is reported on Qwen3-4B-Instruct-2507, since each row requires a separate training run.

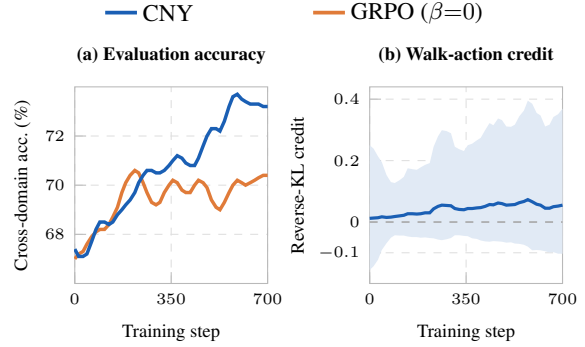
\begin{figure}[t]
\centering
\definecolor{opdblue}{HTML}{1A5FB4}
\definecolor{grpoorange}{HTML}{E07B39}
{\centering\begin{tikzpicture}
\draw[opdblue,line width=1.4pt](0,0)--(0.42,0);\node[anchor=west,font=\footnotesize,inner sep=2pt]at(0.42,0){CNY};
\draw[grpoorange,line width=1.4pt](2.75,0)--(3.17,0);\node[anchor=west,font=\footnotesize,inner sep=2pt]at(3.17,0){GRPO ($\beta{=}0$)};
\end{tikzpicture}\par}
\vspace{0.2em}
\begin{tikzpicture}
\begin{groupplot}[cnyaxis,
  group style={group size=2 by 1, horizontal sep=1.35cm},
  width=2.55cm, height=2.6cm, scale only axis,
  label style={font=\scriptsize}, tick label style={font=\tiny},
  title style={at={(0.5,1)}, anchor=south, yshift=-1pt, font=\scriptsize\bfseries},
  grid=major, ylabel style={yshift=-2pt},
  every axis plot/.append style={line width=1.0pt}]
\nextgroupplot[title={(a) Evaluation accuracy}, xlabel={Training step}, ylabel={Cross-domain acc.\ (\%)},
  xmin=0,xmax=700, xtick={0,350,700}, ymin=66,ymax=74, ytick={68,70,72}]
\addplot[grpoorange, line width=1.3pt] coordinates {(0,67.0) (14,67.2) (30,67.3) (44,67.6) (61,67.9) (77,68.1) (91,68.2) (108,68.2) (124,68.4) (138,68.7) (155,69.1) (171,69.7) (185,70.1) (201,70.4) (218,70.6) (232,70.5) (248,70.1) (262,69.7) (279,69.3) (295,69.2) (309,69.3) (325,69.7) (342,70.0) (356,70.2) (372,70.1) (389,69.8) (403,69.7) (419,69.7) (435,70.0) (449,70.2) (466,70.1) (480,69.9) (496,69.4) (513,69.1) (527,69.0) (543,69.3) (560,69.7) (574,70.0) (590,70.2) (606,70.1) (620,70.0) (637,70.1) (653,70.2) (667,70.3) (684,70.4) (700,70.4)};
\addplot[opdblue, line width=1.3pt] coordinates {(0,67.4) (14,67.1) (30,67.1) (44,67.2) (61,67.7) (77,68.2) (91,68.5) (108,68.5) (124,68.4) (138,68.5) (155,68.8) (171,69.0) (185,69.2) (201,69.4) (218,69.7) (232,70.1) (248,70.4) (262,70.6) (279,70.6) (295,70.5) (309,70.5) (325,70.6) (342,70.8) (356,71.0) (372,71.2) (389,71.1) (403,70.9) (419,70.8) (435,70.8) (449,71.1) (466,71.6) (480,72.0) (496,72.3) (513,72.3) (527,72.2) (543,72.6) (560,73.2) (574,73.6) (590,73.7) (606,73.5) (620,73.4) (637,73.3) (653,73.3) (667,73.3) (684,73.2) (700,73.2)};
\nextgroupplot[title={(b) Walk-action credit}, xlabel={Training step}, ylabel={Reverse-KL credit},
  xmin=0,xmax=700, xtick={0,350,700}, ymin=-0.20,ymax=0.44, ytick={-0.1,0,0.2,0.4}]
\addplot[fill=opdblue!13, draw=none, forget plot] coordinates {(0,0.250) (14,0.239) (30,0.214) (44,0.195) (61,0.153) (77,0.130) (91,0.127) (108,0.136) (124,0.147) (138,0.175) (155,0.172) (171,0.170) (185,0.178) (201,0.186) (218,0.192) (232,0.245) (248,0.279) (262,0.299) (279,0.296) (295,0.289) (309,0.245) (325,0.237) (342,0.231) (356,0.250) (372,0.255) (389,0.266) (403,0.274) (419,0.299) (435,0.322) (449,0.315) (466,0.337) (480,0.332) (496,0.311) (513,0.310) (527,0.317) (543,0.344) (560,0.363) (574,0.395) (590,0.385) (606,0.355) (620,0.347) (637,0.311) (653,0.317) (667,0.342) (684,0.351) (700,0.369) (700,-0.103) (684,-0.103) (667,-0.098) (653,-0.099) (637,-0.095) (620,-0.092) (606,-0.091) (590,-0.092) (574,-0.080) (560,-0.074) (543,-0.071) (527,-0.070) (513,-0.066) (496,-0.067) (480,-0.064) (466,-0.063) (449,-0.062) (435,-0.064) (419,-0.063) (403,-0.060) (389,-0.055) (372,-0.053) (356,-0.046) (342,-0.045) (325,-0.043) (309,-0.043) (295,-0.052) (279,-0.056) (262,-0.059) (248,-0.057) (232,-0.053) (218,-0.052) (201,-0.050) (185,-0.051) (171,-0.049) (155,-0.049) (138,-0.047) (124,-0.044) (108,-0.044) (91,-0.043) (77,-0.051) (61,-0.069) (44,-0.091) (30,-0.123) (14,-0.144) (0,-0.156)} \closedcycle;
\addplot[gray!70, dashed, line width=0.6pt, forget plot] coordinates {(0,0) (700,0)};
\addplot[opdblue, line width=1.2pt] coordinates {(0,0.012) (14,0.013) (30,0.014) (44,0.017) (61,0.015) (77,0.016) (91,0.018) (108,0.020) (124,0.022) (138,0.027) (155,0.027) (171,0.026) (185,0.027) (201,0.030) (218,0.030) (232,0.043) (248,0.051) (262,0.055) (279,0.054) (295,0.053) (309,0.044) (325,0.041) (342,0.040) (356,0.044) (372,0.044) (389,0.046) (403,0.047) (419,0.051) (435,0.057) (449,0.056) (466,0.062) (480,0.060) (496,0.055) (513,0.056) (527,0.057) (543,0.063) (560,0.067) (574,0.073) (590,0.066) (606,0.058) (620,0.055) (637,0.045) (653,0.045) (667,0.050) (684,0.052) (700,0.055)};
\end{groupplot}
\end{tikzpicture}
\caption{\textbf{Effectiveness of OPSD} (CNY-7B). \textbf{(a)} Held-out accuracy (mean of WikiCS and Cora) over training for $\beta{=}0.03$ vs.\ the $\beta{=}0$ GRPO ablation. \textbf{(b)} OPSD walk-action reverse-KL signal $-\delta_t$ on walk-action tokens; band: $1$st--$99$th percentile, line: mean.}
\label{fig:opd-effectiveness}
\end{figure}

\paragraph{OPSD improves cross-domain transfer.} Figure~\ref{fig:opd-effectiveness}(a) tracks held-out accuracy (mean of WikiCS and Cora) over training for CNY and the GRPO ablation ($\beta{=}0$). Both reach the same training reward, yet CNY attains a markedly higher cross-domain accuracy. Matched training reward alongside a separated evaluation accuracy places the contribution of OPSD in generalisation rather than in a closer fit to the training mixture.

\paragraph{The OPSD credit is a structured, bidirectional signal.} Figure~\ref{fig:opd-effectiveness}(b) shows the per-token reverse KL $\mathrm{KL}^{\mathrm{rev}}_t = -\delta_t$ on the walk-action tokens through training. The signal spans both teacher-favoured selections (below zero, where the recap makes the action more likely) and teacher-disfavoured ones (above zero), confirming a genuine walk-action credit rather than a constant offset. Its $1$st-$99$th percentile spread widens as training proceeds, indicating that the credit sharpens its discrimination between strong and weak selections rather than washing out. Across $7{,}228$ training recaps, $0.0\%$ contain node identifiers or walk recommendations, and only $0.03\%$ (two benign cases) touch the start node’s classification. The mean recap length is $32$ words, below the $50$-token cap (App.~\ref{app:recap-prompt}).

\begin{table}[t]
\centering
\small
\setlength{\tabcolsep}{5pt}
\begin{tabular}{lccccc}
\toprule
                 & \multicolumn{5}{c}{OPSD coefficient $\beta$} \\
\cmidrule(lr){2-6}
Dataset          & $0$ (GRPO) & $0.005$ & $0.01$ & $0.02$ & $\mathbf{0.03}$ \\
\midrule
WikiCS           & $69.5$ & $72.7$ & $73.0$ & $74.2$ & $\mathbf{75.1}$ \\
Cora             & $58.5$ & $60.0$ & $62.5$ & $62.0$ & $\mathbf{65.5}$ \\
\bottomrule
\end{tabular}

\caption{\textbf{$\beta$-sweep, held-out accuracy} (Qwen3-4B-Instruct-2507). $\beta{=}0$ is the GRPO ablation. \textbf{Bold}: best per row.}
\label{tab:beta-sweep}
\end{table}

\paragraph{The improvement scales with OPSD strength.} On a smaller Qwen3-4B-Instruct-2507 backbone, sweeping $\beta$ from 0 (GRPO) up to the headline 0.03 improves held-out accuracy on both datasets (Table~\ref{tab:beta-sweep}): WikiCS rises monotonically from $69.5$ to $75.1$ ($+5.6$), and Cora follows the same upward trend from $58.5$ to $65.5$ ($+7.0$). This single coefficient also transfers. Measured on frozen checkpoints (Table~\ref{tab:credit-stats}), the recap raises the taken walk's probability on $77$--$89\%$ of walk actions at every scale, and although the signal strength grows with backbone capacity, $\beta{=}0.03$ keeps the per-token nudge well below reward scale.

\begin{table}[t]
\centering
\footnotesize
\setlength{\tabcolsep}{4pt}
\begin{tabular}{lccc}
\toprule
Frozen checkpoint & mean $|\delta|$ & frac $\delta{>}0$ & mean $|\beta\delta|$ \\
\midrule
Qwen3-4B (sweep)      & $0.45$ & $79\%$ & $0.013$ \\
Qwen2.5-7B            & $0.56$ & $77\%$ & $0.017$ \\
Qwen2.5-14B (main)    & $0.91$ & $89\%$ & $0.027$ \\
\bottomrule
\end{tabular}
\caption{\textbf{The OPSD signal across backbone scales}, in nats over the \texttt{<walk>} id tokens (Cora, $N{=}536$ per backbone). The credit strengthens with capacity while $\beta{=}0.03$ keeps its per-token magnitude well below reward scale.}
\label{tab:credit-stats}
\end{table}

\paragraph{The reward tiers are not the source.} Table~\ref{tab:reward-ablation} ablates the graded reward of Eq.~\ref{eq:reward-tier} at identical data and rollouts. Rescaling every tier value reproduces both datasets, while zeroing the two intermediate walk tiers collapses rollout behaviour, so tier ordering matters and the specific values do not.

\begin{table}[t]
\centering
\footnotesize
\setlength{\tabcolsep}{3.5pt}
\begin{tabular}{lccc}
\toprule
                  & Full & Rescaled & Removed \\
\midrule
WikiCS            & $79.2${\scriptsize$\pm1.2$} & $79.2${\scriptsize$\pm0.8$} & \tval \\
Cora              & $61.9${\scriptsize$\pm1.3$} & $62.9${\scriptsize$\pm1.4$} & \tval \\
Walk attempts/rollout & $2.3$   & $2.0$   & $2.3\!\to\!13.7$ \\
Budget-exhaust.   & $0\%$   & $0.4\%$ & $0.4\!\to\!75\%$ \\
Answer rate       & $100\%$ & $98\%$  & $99.6\!\to\!25\%$ \\
\bottomrule
\end{tabular}
\caption{\textbf{Component-wise reward ablation} (Qwen3-4B, $n{=}200$ held-out, matched data and rollouts). Tiers: $1.0/0.6/0.3/0.2/0.1$ (Full, Eq.~\ref{eq:reward-tier}), order-preserving $1.0/0.5/0.25/0.15/0.05$ (Rescaled), $1.0/0.6/0/0/0.1$ (Removed, zeroing the two walk tiers for incorrect trajectories). Accuracy is unavailable once the policy stops answering.}
\label{tab:reward-ablation}
\end{table}
 \textbf{Together these place the OPSD gain in generalisation rather than in reward shaping or a fragile coefficient choice.}

\subsection{Impact of Different LLM Backbones}
\label{sec:capability-gap}

\begin{figure}[t]
\centering
\definecolor{cnyHero}{HTML}{7D3C52}
\definecolor{cnyCora}{HTML}{2F6E8F}
\begin{tikzpicture}
\begin{axis}[cnyaxis, width=6.4cm, height=3.0cm, scale only axis,
  xtick={1,2,3,4,5}, xmin=0.5, xmax=5.5,
  xticklabels={Llama-3.2\\3B, Qwen3\\4B, Qwen2.5\\7B, Qwen2.5\\14B, Qwen3\\14B},
  xticklabel style={font=\footnotesize, align=center, yshift=1pt},
  ymin=32, ymax=88, ytick={40,60,80}, ymajorgrids,
  ylabel={Accuracy (\%)}, ylabel style={yshift=-3pt},
  label style={font=\small}, tick label style={font=\footnotesize},
  legend style={at={(0.985,0.04)}, anchor=south east, legend columns=2, font=\scriptsize,
    draw=gray!40, fill=white, fill opacity=0.92, text opacity=1, nodes={inner sep=1.3pt},
    /tikz/every even column/.append style={column sep=4pt}}]
\draw[cnyHero!38, line width=1.7pt, line cap=round] (axis cs:0.86,48.0)--(axis cs:0.86,71.5);
\draw[cnyHero!38, line width=1.7pt, line cap=round] (axis cs:1.86,76.0)--(axis cs:1.86,83.5);
\draw[cnyHero!38, line width=1.7pt, line cap=round] (axis cs:2.86,68.8)--(axis cs:2.86,76.6);
\draw[cnyHero!38, line width=1.7pt, line cap=round] (axis cs:3.86,74.0)--(axis cs:3.86,79.2);
\draw[cnyHero!38, line width=1.7pt, line cap=round] (axis cs:4.86,74.6)--(axis cs:4.86,79.6);
\draw[cnyCora!38, line width=1.7pt, line cap=round] (axis cs:1.14,36.0)--(axis cs:1.14,68.0);
\draw[cnyCora!38, line width=1.7pt, line cap=round] (axis cs:2.14,60.5)--(axis cs:2.14,69.5);
\draw[cnyCora!38, line width=1.7pt, line cap=round] (axis cs:3.14,66.8)--(axis cs:3.14,72.4);
\draw[cnyCora!38, line width=1.7pt, line cap=round] (axis cs:4.14,64.4)--(axis cs:4.14,71.4);
\draw[cnyCora!38, line width=1.7pt, line cap=round] (axis cs:5.14,60.6)--(axis cs:5.14,66.4);
\addplot[only marks, mark=*, mark size=2.3pt, forget plot, mark options={fill=white, draw=cnyHero, line width=0.8pt}]
  coordinates {(0.86,48.0)(1.86,76.0)(2.86,68.8)(3.86,74.0)(4.86,74.6)};
\addplot[only marks, mark=*, mark size=2.5pt, forget plot, mark options={fill=cnyHero, draw=cnyHero!55!black}]
  coordinates {(0.86,71.5)(1.86,83.5)(2.86,76.6)(3.86,79.2)(4.86,79.6)};
\addplot[only marks, mark=*, mark size=2.3pt, forget plot, mark options={fill=white, draw=cnyCora, line width=0.8pt}]
  coordinates {(1.14,36.0)(2.14,60.5)(3.14,66.8)(4.14,64.4)(5.14,60.6)};
\addplot[only marks, mark=*, mark size=2.5pt, forget plot, mark options={fill=cnyCora, draw=cnyCora!55!black}]
  coordinates {(1.14,68.0)(2.14,69.5)(3.14,72.4)(4.14,71.4)(5.14,66.4)};
\addlegendimage{only marks, mark=*, mark size=2.3pt, mark options={fill=white, draw=cnyHero, line width=0.8pt}}
\addlegendentry{WikiCS (Base)}
\addlegendimage{only marks, mark=*, mark size=2.5pt, mark options={fill=cnyHero, draw=cnyHero!55!black}}
\addlegendentry{WikiCS (CNY)}
\addlegendimage{only marks, mark=*, mark size=2.3pt, mark options={fill=white, draw=cnyCora, line width=0.8pt}}
\addlegendentry{Cora (Base)}
\addlegendimage{only marks, mark=*, mark size=2.5pt, mark options={fill=cnyCora, draw=cnyCora!55!black}}
\addlegendentry{Cora (CNY)}
\end{axis}
\end{tikzpicture}
\caption{\textbf{CNY improves accuracy across base models.} Zero-shot base (hollow) vs.\ CNY best checkpoint (filled) for five backbones on WikiCS (maroon) and Cora (blue). Accuracy is on the evaluation subset tracked during training, so values differ slightly from Table~\ref{tab:main}.}
\label{fig:base-lift}
\end{figure}
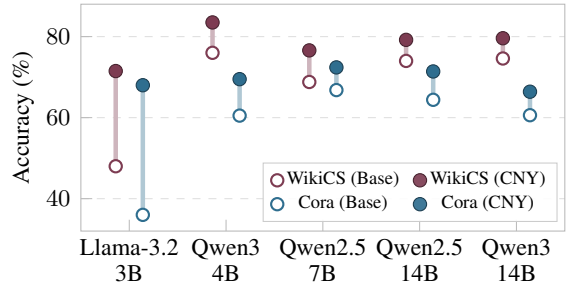

To assess the generality of CNY across LLMs, the method is trained from five base models spanning two families and four scales (Llama-3.2-3B and Qwen 4B--14B), with reward, OPSD coefficient and training mixture held fixed. Held-out accuracy on the evaluation subset tracked during training is reported in Fig.~\ref{fig:base-lift}.

On both evaluation datasets the accuracy of every backbone improves after CNY training, the largest gain falling to the weakest base and the improvement shrinking as the backbone strengthens. \textbf{The improvement is therefore a property of CNY rather than of any single backbone or dataset.}

\subsection{Computation Cost}
\label{sec:analysis}

CNY's only algorithmic overhead beyond GRPO is one teacher log-probability forward per successful $\texttt{<walk>}$. Because this forward operates on a bounded-length recap rather than retrieved neighbour text (App.~\ref{app:recap-prompt}), its cost is independent of the graph context size. On a matched Qwen2.5-7B setup, OPSD increases log-probability time from $14.1$ to $17.6$\,s/step and aggregate compute from $12.0$ to $14.8$\,PFLOP (+24\%; Table~\ref{tab:cost}). The remaining wall-clock overhead comes from the longer trajectories induced by learned walking.

At inference, per-node cost depends only on the neighbourhood the walk exposes and the step budget. Cost therefore scales with the explored local context, not the global graph size.

\begin{table}[t]
\centering
\small
\setlength{\tabcolsep}{4pt}
\resizebox{\columnwidth}{!}{%
\begin{tabular}{l c c}
\toprule
Per-step phase (s)            & GRPO ($\beta{=}0$) & CNY{+}OPSD \\
\midrule
Rollout                       & $8.8$              & $20.0$ \\
Log probabilities             & $14.1$             & $17.6$ \\
Actor train (fwd/bwd)         & $48.3$             & $55.3$ \\
\midrule
Total step wall               & $87.7$             & $108.9$ \\
Compute (PFLOP/step)          & $12.0$             & $14.8$ \\
\bottomrule
\end{tabular}%
}

\caption{\textbf{Training-time cost.} Median per-step wall time and compute for a matched 7B pair on identical data and hardware, with and without OPSD.}
\label{tab:cost}
\end{table}

\subsection{Case Study}
\label{sec:case-study}

A representative WikiCS case illustrates how walking can resolve a direct misclassification. For the target node with ground-truth label \texttt{Distributed Computing Architecture}, the frozen CNY-14B model predicts \texttt{Computer Architecture} even when given two complete 1-hop neighbours. With walking enabled, the model selects the definitional \texttt{cloud\_computing} article rather than the less discriminative \texttt{rackspace} company page, leading to the correct prediction. Across the WikiCS analysis set, walking corrects 290 direct errors while introducing 91 new errors. \begin{tcolorbox}[
  enhanced, breakable,
  colback=promptbg, colframe=tagwrong,
  boxrule=0.6pt, arc=2.5pt, left=4pt, right=4pt, top=2pt, bottom=2pt,
  title={No walk $\rightarrow$ Computer Architecture\ \ (incorrect)},
  fonttitle=\bfseries\footnotesize, coltitle=white]
\begin{lstlisting}[style=cnyprompt]
<thinking>
The target article discusses Quanta Cloud Technology (QCT), which provides data center hardware and cloud solutions ... the most fitting category would be related to the physical and architectural aspects of computing systems, which aligns with Computer Architecture.
</thinking>
<answer>3</answer>
\end{lstlisting}
\end{tcolorbox}

\vspace{2pt}

\begin{tcolorbox}[
  enhanced, breakable,
  colback=promptbg, colframe=taganswer,
  boxrule=0.6pt, arc=2.5pt, left=4pt, right=4pt, top=2pt, bottom=2pt,
  title={Walk $\rightarrow$ Distributed Computing Architecture\ \ (correct)},
  fonttitle=\bfseries\footnotesize, coltitle=white]
\begin{lstlisting}[style=cnyprompt]
<thinking>The target article is about Quanta Cloud Technology (QCT), which deals with cloud solutions ... I need to walk to node 4194 to understand more about cloud computing and confirm the category.</thinking>

<walk>4194</walk>

<information> [#4194 cloud_computing] cloud computing on-demand availability computer system resources especially data storage computing power ... large clouds predominant today often functions distributed multiple locations central servers ... clouds may limited single organization enterprise clouds available many organizations public cloud ... cloud computing relies sharing resources achieve coherence economies scale ... [1-hop neighbours elided] </information>

<thinking>This confirms the context is related to cloud computing, which falls under Distributed Computing Architecture.</thinking>

<answer>7</answer>
\end{lstlisting}
\end{tcolorbox}

The introduced errors fall into three failure modes based on all 282 flips pooled across three walk-evaluation seeds: \textbf{adjacent-class interference} (53.5\%), where a semantically adjacent neighbour dominates the reasoning; \textbf{misleading emphasis} (39.4\%), where a same-class neighbour foregrounds an off-class facet; and \textbf{reasoning drift} (6.7\%), where the final prediction matches neither the target nor any visited neighbour. Over 92\% thus arise from plausible but insufficiently discriminative evidence rather than search failure, and no case exhausts the walk budget (App.~\ref{app:failure-modes}).
\section{Conclusion}

This paper introduces CNY, a reinforcement learning framework that treats neighbour acquisition as graph-walk actions and supervises them via destination-conditioned on-policy self-distillation (OPSD), which derives action-level credit from revealed destinations without annotated trajectories, external judges, or additional rollouts. Across multiple TAG benchmarks, CNY consistently improves reasoning performance and transfers to unseen domains, suggesting that effective graph reasoning depends not only on interpreting evidence but also on acquiring it adaptively during reasoning.

\section*{Limitations}

CNY trains walk selection rather than the backbone's underlying world knowledge, so final accuracy is bounded by what the base model already encodes and training the walk LLM cannot inject domain facts the backbone never saw. The base-model study (\S\ref{sec:capability-gap}) reflects this, with held-out accuracy tracking backbone scale and the gain from walking shrinking as the backbone strengthens. Tasks that hinge on knowledge absent from the base model are unlikely to be reached without a stronger or further pre-trained backbone.

Evaluation uses the GOFA-aligned zero-shot benchmark inherited from Graph-R1 and TRN-R1-Zero, which keeps every comparison directly aligned with prior work but does not cover large-scale, heterogeneous or temporal graphs. The cross-task generalisation evidence further rests on one graph-level task (Expla-Graph) and one open-ended question-answering task (WebQSP), so broader graph families and task types are left to future work.

The backbone study spans the Llama and Qwen families up to 14B, leaving other model families uncharacterised.

\section*{Acknowledgments}

This research has been supported by Australian Research Council Discovery Projects (CE200100025, DP230101196 and DE250100919).

\bibliography{custom}

@inproceedings{wu2025graphr1,
  author    = {Yicong Wu and
               Guangyue Lu and
               Yuan Zuo and
               Huarong Zhang and
               Junjie Wu},
  title     = {{Graph-R1}: Incentivizing the Zero-Shot Graph Learning Capability in
               {LLMs} via Explicit Reasoning},
  booktitle = {{{EMNLP}}},
  year      = {2025},
}

@inproceedings{liu2026trnr1zero,
  author       = {Yilun Liu and
                  Ruihong Qiu and
                  Zi Huang},
  title        = {{TRN-R1-Zero}: Text-rich Network Reasoning via {LLMs} with Reinforcement
                  Learning Only},
  booktitle    = {ACL},
  year         = {2026},
}

@article{schulman2017ppo,
  author    = {John Schulman and
               Filip Wolski and
               Prafulla Dhariwal and
               Alec Radford and
               Oleg Klimov},
  title     = {Proximal Policy Optimization Algorithms},
  journal   = {CoRR},
  volume    = {abs/1707.06347},
  year      = {2017},
}

@inproceedings{hubotter2026sdpo,
  author    = {Jonas H{\"u}botter and
               Frederike L{\"u}beck and
               Lejs Behric and
               Anton Baumann and
               Marco Bagatella and
               Daniel Marta and
               Ido Hakimi and
               Idan Shenfeld and
               Thomas Kleine Buening and
               Carlos Guestrin and
               Andreas Krause},
  title     = {Reinforcement Learning via Self-Distillation},
  booktitle = {{{ICML}}},
  year      = {2026},
}

@inproceedings{zhao2026opsd,
  author    = {Siyan Zhao and
               Zhihui Xie and
               Mengchen Liu and
               Jing Huang and
               Guan Pang and
               Feiyu Chen and
               Aditya Grover},
  title     = {Self-Distilled Reasoner: On-Policy Self-Distillation for Large Language Models},
  booktitle = {{{ICML}}},
  year      = {2026},
}

@article{kim2026whyselfdistill,
  author    = {Jeonghye Kim and
               Xufang Luo and
               Minbeom Kim and
               Sangmook Lee and
               Dohyung Kim and
               Jiwon Jeon and
               Dongsheng Li and
               Yuqing Yang},
  title     = {Why Does Self-Distillation (Sometimes) Degrade the Reasoning
               Capability of {LLMs}?},
  journal   = {CoRR},
  volume    = {abs/2603.24472},
  year      = {2026},
}

@article{wang2026openclaw,
  author    = {Yinjie Wang and
               Xuyang Chen and
               Xiaolong Jin and
               Mengdi Wang and
               Ling Yang},
  title     = {OpenClaw-RL: Train Any Agent Simply by Talking},
  journal   = {CoRR},
  volume    = {abs/2603.10165},
  year      = {2026},
}

@article{shenfeld2026sdft,
  author    = {Idan Shenfeld and
               Mehul Damani and
               Jonas H{\"u}botter and
               Pulkit Agrawal},
  title     = {Self-Distillation Enables Continual Learning},
  journal   = {CoRR},
  volume    = {abs/2601.19897},
  year      = {2026},
}

@article{shao2024deepseekmath,
  author    = {Zhihong Shao and
               Peiyi Wang and
               Qihao Zhu and
               Runxin Xu and
               Junxiao Song and
               Mingchuan Zhang and
               Y. K. Li and
               Y. Wu and
               Daya Guo},
  title     = {{DeepSeekMath}: Pushing the Limits of Mathematical Reasoning in Open
               Language Models},
  journal   = {CoRR},
  volume    = {abs/2402.03300},
  year      = {2024},
}

@inproceedings{lightman2024prm800k,
  author    = {Hunter Lightman and
               Vineet Kosaraju and
               Yuri Burda and
               Harrison Edwards and
               Bowen Baker and
               Teddy Lee and
               Jan Leike and
               John Schulman and
               Ilya Sutskever and
               Karl Cobbe},
  title     = {Let's Verify Step by Step},
  booktitle = {{{ICLR}}},
  year      = {2024},
}

@inproceedings{wang2024mathshepherd,
  author    = {Peiyi Wang and
               Lei Li and
               Zhihong Shao and
               Runxin Xu and
               Damai Dai and
               Yifei Li and
               Deli Chen and
               Yu Wu and
               Zhifang Sui},
  title     = {{Math-Shepherd}: Verify and Reinforce {LLMs} Step-by-step without Human
               Annotations},
  booktitle = {{{ACL}}},
  year      = {2024},
}

@inproceedings{li2024zerog,
  author    = {Yuhan Li and
               Peisong Wang and
               Zhixun Li and
               Jeffrey Xu Yu and
               Jia Li},
  title     = {{ZeroG}: Investigating Cross-dataset Zero-shot Transferability in Graphs},
  booktitle = {{{KDD}}},
  year      = {2024},
}

@inproceedings{he2024tape,
  author    = {Xiaoxin He and
               Xavier Bresson and
               Thomas Laurent and
               Adam Perold and
               Yann LeCun and
               Bryan Hooi},
  title     = {Harnessing Explanations: {LLM-to-LM} Interpreter for Enhanced Text-Attributed
               Graph Representation Learning},
  booktitle = {{{ICLR}}},
  year      = {2024},
}

@inproceedings{tang2024graphgpt,
  author    = {Jiabin Tang and
               Yuhao Yang and
               Wei Wei and
               Lei Shi and
               Lixin Su and
               Suqi Cheng and
               Dawei Yin and
               Chao Huang},
  title     = {{GraphGPT}: Graph Instruction Tuning for Large Language Models},
  booktitle = {{{SIGIR}}},
  year      = {2024},
}

@inproceedings{chen2024llaga,
  author    = {Runjin Chen and
               Tong Zhao and
               Ajay Kumar Jaiswal and
               Neil Shah and
               Zhangyang Wang},
  title     = {{LLaGA}: Large Language and Graph Assistant},
  booktitle = {{{ICML}}},
  year      = {2024},
}

@inproceedings{kong2025gofa,
  author    = {Lecheng Kong and
               Jiarui Feng and
               Hao Liu and
               Chengsong Huang and
               Jiaxin Huang and
               Yixin Chen and
               Muhan Zhang},
  title     = {{GOFA:} {A} Generative One-For-All Model for Joint Graph Language
               Modeling},
  booktitle = {{{ICLR}}},
  year      = {2025},
}

@article{touvron2023llama2,
  author    = {Hugo Touvron and
               Louis Martin and
               Kevin Stone and
               Peter Albert and
               Amjad Almahairi and
               Yasmine Babaei and
               Nikolay Bashlykov and
               Soumya Batra and
               Prajjwal Bhargava and
               Shruti Bhosale and
               Dan Bikel and
               Lukas Blecher and
               Cristian Canton{-}Ferrer and
               Moya Chen and
               Guillem Cucurull and
               David Esiobu and
               Jude Fernandes and
               Jeremy Fu and
               Wenyin Fu and
               others},
  title     = {{Llama 2}: Open Foundation and Fine-Tuned Chat Models},
  journal   = {CoRR},
  volume    = {abs/2307.09288},
  year      = {2023},
}

@article{jiang2023mistral,
  author    = {Albert Q. Jiang and
               Alexandre Sablayrolles and
               Arthur Mensch and
               Chris Bamford and
               Devendra Singh Chaplot and
               Diego de Las Casas and
               Florian Bressand and
               Gianna Lengyel and
               Guillaume Lample and
               Lucile Saulnier and
               L{\'{e}}lio Renard Lavaud and
               Marie{-}Anne Lachaux and
               Pierre Stock and
               Teven Le Scao and
               Thibaut Lavril and
               Thomas Wang and
               Timoth{\'{e}}e Lacroix and
               William El Sayed},
  title     = {{Mistral 7B}},
  journal   = {CoRR},
  volume    = {abs/2310.06825},
  year      = {2023},
}

@inproceedings{liu2024oneforall,
  author    = {Hao Liu and
               Jiarui Feng and
               Lecheng Kong and
               Ningyue Liang and
               Dacheng Tao and
               Yixin Chen and
               Muhan Zhang},
  title     = {One For All: Towards Training One Graph Model For All Classification
               Tasks},
  booktitle = {{{ICLR}}},
  year      = {2024},
}

@inproceedings{he2025unigraph,
  author    = {Yufei He and
               Yuan Sui and
               Xiaoxin He and
               Bryan Hooi},
  title     = {{UniGraph}: Learning a Unified Cross-Domain Foundation Model for Text-Attributed
               Graphs},
  booktitle = {{{KDD}}},
  year      = {2025},
}

@inproceedings{chen2024graphwiz,
  author    = {Nuo Chen and
               Yuhan Li and
               Jianheng Tang and
               Jia Li},
  title     = {{GraphWiz}: An Instruction-Following Language Model for Graph Computational
               Problems},
  booktitle = {{{KDD}}},
  year      = {2024},
}

@inproceedings{wu2025llmnodebed,
  author    = {Xixi Wu and
               Yifei Shen and
               Fangzhou Ge and
               Caihua Shan and
               Yizhu Jiao and
               Xiangguo Sun and
               Hong Cheng},
  title     = {When Do {LLMs} Help With Node Classification? {A} Comprehensive Analysis},
  booktitle = {{{ICML}}},
  year      = {2025},
}

@article{wang2025npgmuse,
  author    = {Yuyao Wang and Bowen Liu and Jianheng Tang and Nuo Chen and Yuhan Li and Qifan Zhang and Chenyi Zi and Chen Zhang and Jia Li},
  title     = {NPG-Muse: Scaling Long Chain-of-Thought Reasoning with NP-Hard Graph Problems},
  journal   = {CoRR},
  volume    = {abs/2508.20373},
  year      = {2025},
}

@inproceedings{jin2025searchr1,
  author    = {Bowen Jin and
               Hansi Zeng and
               Zhenrui Yue and
               Jinsung Yoon and
               Sercan Arik and
               Dong Wang and
               Hamed Zamani and
               Jiawei Han},
  title     = {{Search-R1}: Training {LLMs} to Reason and Leverage Search Engines with Reinforcement Learning},
  booktitle = {{{COLM}}},
  year      = {2025},
}

@article{song2025r1searcher,
  author    = {Huatong Song and
               Jinhao Jiang and
               Yingqian Min and
               Jie Chen and
               Zhipeng Chen and
               Wayne Xin Zhao and
               Lei Fang and
               Ji-Rong Wen},
  title     = {{R1-Searcher}: Incentivizing the Search Capability in {LLMs} via Reinforcement
               Learning},
  journal   = {CoRR},
  volume    = {abs/2503.05592},
  year      = {2025},
}

@article{qwen25,
  author    = {An Yang and
               Baosong Yang and
               Beichen Zhang and
               Binyuan Hui and
               Bo Zheng and
               Bowen Yu and
               Chengyuan Li and
               Dayiheng Liu and
               Fei Huang and
               Haoran Wei and
               Huan Lin and
               Jian Yang and
               Jianhong Tu and
               Jianwei Zhang and
               Jianxin Yang and
               Jiaxi Yang and
               Jingren Zhou and
               Junyang Lin and
               Kai Dang and
               Keming Lu and
               Keqin Bao and
               Kexin Yang and
               Le Yu and
               Mei Li and
               Mingfeng Xue and
               Pei Zhang and
               Qin Zhu and
               Rui Men and
               Runji Lin and
               Tianhao Li and
               Tingyu Xia and
               Xingzhang Ren and
               Xuancheng Ren and
               Yang Fan and
               Yang Su and
               Yichang Zhang and
               Yu Wan and
               Yuqiong Liu and
               Zeyu Cui and
               Zhenru Zhang and
               Zihan Qiu},
  title     = {{Qwen2.5} Technical Report},
  journal   = {CoRR},
  volume    = {abs/2412.15115},
  year      = {2024},
}

@article{liu2024lostmiddle,
  author    = {Nelson F. Liu and
               Kevin Lin and
               John Hewitt and
               Ashwin Paranjape and
               Michele Bevilacqua and
               Fabio Petroni and
               Percy Liang},
  title     = {Lost in the Middle: How Language Models Use Long Contexts},
  journal   = {TACL},
  year      = {2024},
}

@inproceedings{hsieh2024ruler,
  author    = {Cheng-Ping Hsieh and
               Simeng Sun and
               Samuel Kriman and
               Shantanu Acharya and
               Dima Rekesh and
               Fei Jia and
               Yang Zhang and
               Boris Ginsburg},
  title     = {{RULER}: What's the Real Context Size of Your Long-Context Language Models?},
  booktitle = {{{COLM}}},
  year      = {2024},
}

@inproceedings{bai2024longbench,
  author    = {Yushi Bai and
               Xin Lv and
               Jiajie Zhang and
               Hongchang Lyu and
               Jiankai Tang and
               Zhidian Huang and
               Zhengxiao Du and
               Xiao Liu and
               Aohan Zeng and
               Lei Hou and
               Yuxiao Dong and
               Jie Tang and
               Juanzi Li},
  title     = {{LongBench}: {A} Bilingual, Multitask Benchmark for Long Context Understanding},
  booktitle = {{{ACL}}},
  year      = {2024},
}

@article{drgrpo,
  author       = {Zichen Liu and
                  Changyu Chen and
                  Wenjun Li and
                  Penghui Qi and
                  Tianyu Pang and
                  Chao Du and
                  Wee Sun Lee and
                  Min Lin},
  title        = {Understanding R1-Zero-Like Training: {A} Critical Perspective},
  journal      = {CoRR},
  volume       = {abs/2503.20783},
  year         = {2025},
}

@article{guo2025deepseekr1,
  author    = {Daya Guo and
               Dejian Yang and
               Haowei Zhang and
               Junxiao Song and
               Peiyi Wang and
               Qihao Zhu and
               Runxin Xu and
               Ruoyu Zhang and
               Shirong Ma and
               Xiao Bi and others},
  title     = {{DeepSeek-R1} Incentivizes Reasoning in {LLMs} through Reinforcement Learning},
  journal   = {Nature},
  volume    = {645},
  number    = {8081},
  pages     = {633--638},
  year      = {2025},
}

@inproceedings{wei2022cot,
  author    = {Jason Wei and
               Xuezhi Wang and
               Dale Schuurmans and
               Maarten Bosma and
               Brian Ichter and
               Fei Xia and
               Ed H. Chi and
               Quoc V. Le and
               Denny Zhou},
  title     = {Chain-of-Thought Prompting Elicits Reasoning in Large Language Models},
  booktitle = {{{NeurIPS}}},
  year      = {2022},
}

@article{llamateam2024llama3,
  author    = {{Llama Team}},
  title     = {The {Llama 3} Herd of Models},
  journal   = {CoRR},
  volume    = {abs/2407.21783},
  year      = {2024},
}

@inproceedings{li2024glbench,
  author    = {Yuhan Li and
               Peisong Wang and
               Xiao Zhu and
               Aochuan Chen and
               Haiyun Jiang and
               Deng Cai and
               Wai Kin Victor Chan and
               Jia Li},
  title     = {{GLBench}: {A} Comprehensive Benchmark for Graph with Large Language Models},
  booktitle = {{{NeurIPS}}},
  year      = {2024},
}

@inproceedings{wang2023nlgraph,
  author    = {Heng Wang and
               Shangbin Feng and
               Tianxing He and
               Zhaoxuan Tan and
               Xiaochuang Han and
               Yulia Tsvetkov},
  title     = {Can Language Models Solve Graph Problems in Natural Language?},
  booktitle = {{{NeurIPS}}},
  year      = {2023},
}

@article{mernyei2020wikics,
  author    = {P{\'{e}}ter Mernyei and
               Catalina Cangea},
  title     = {{Wiki-CS}: {A} Wikipedia-Based Benchmark for Graph Neural Networks},
  journal   = {CoRR},
  volume    = {abs/2007.02901},
  year      = {2020},
}

@inproceedings{hu2020ogb,
  author    = {Weihua Hu and
               Matthias Fey and
               Marinka Zitnik and
               Yuxiao Dong and
               Hongyu Ren and
               Bowen Liu and
               Michele Catasta and
               Jure Leskovec},
  title     = {Open Graph Benchmark: Datasets for Machine Learning on Graphs},
  booktitle = {{{NeurIPS}}},
  year      = {2020},
}

@article{sen2008cora,
  author    = {Prithviraj Sen and
               Galileo Namata and
               Mustafa Bilgic and
               Lise Getoor and
               Brian Gallagher and
               Tina Eliassi-Rad},
  title     = {Collective Classification in Network Data},
  journal   = {{AI} Magazine},
  volume    = {29},
  number    = {3},
  pages     = {93--106},
  year      = {2008},
}

@inproceedings{wang2025llmbp,
  author    = {Haoyu Peter Wang and
               Shikun Liu and
               Rongzhe Wei and
               Pan Li},
  title     = {Generalization Principles for Inference over Text-Attributed Graphs with Large Language Models},
  booktitle = {{{ICML}}},
  year      = {2025},
}

@inproceedings{li2025searcho1,
  author    = {Xiaoxi Li and
               Guanting Dong and
               Jiajie Jin and
               Yuyao Zhang and
               Yujia Zhou and
               Yutao Zhu and
               Peitian Zhang and
               Zhicheng Dou},
  title     = {{Search-o1}: Agentic Search-Enhanced Large Reasoning Models},
  booktitle = {{{EMNLP}}},
  year      = {2025},
}

@inproceedings{asai2024selfrag,
  author    = {Akari Asai and
               Zeqiu Wu and
               Yizhong Wang and
               Avirup Sil and
               Hannaneh Hajishirzi},
  title     = {{Self-RAG}: Learning to Retrieve, Generate, and Critique through Self-Reflection},
  booktitle = {{{ICLR}}},
  year      = {2024},
}

@inproceedings{chen2025research,
  author    = {Mingyang Chen and
               Linzhuang Sun and
               Tianpeng Li and
               Haoze Sun and
               Yijie Zhou and
               Chenzheng Zhu and
               Haofen Wang and
               Jeff Z. Pan and
               Wen Zhang and
               Huajun Chen and
               Fan Yang and
               Zenan Zhou and
               Weipeng Chen},
  title     = {{ReSearch}: Learning to Reason with Search for {LLMs} via Reinforcement Learning},
  booktitle = {{{NeurIPS}}},
  year      = {2025},
}

@inproceedings{yih2016webqsp,
  author    = {Wen{-}tau Yih and
               Matthew Richardson and
               Christopher Meek and
               Ming{-}Wei Chang and
               Jina Suh},
  title     = {The Value of Semantic Parse Labeling for Knowledge Base Question Answering},
  booktitle = {{ACL}},
  year      = {2016},
}

@inproceedings{he2024gretriever,
  author    = {Xiaoxin He and
               Yijun Tian and
               Yifei Sun and
               Nitesh V. Chawla and
               Thomas Laurent and
               Yann LeCun and
               Xavier Bresson and
               Bryan Hooi},
  title     = {{G-Retriever}: Retrieval-Augmented Generation for Textual Graph Understanding and Question Answering},
  booktitle = {{{NeurIPS}}},
  year      = {2024},
}

@article{lexa,
  author       = {Yanran Tang and
                  Ruihong Qiu and
                  Yilun Liu and
                  Xue Li and
                  Zi Huang},
  title        = {{LEXA:} Legal case retrieval via graph contrastive learning with contextualised {LLM} embeddings},
  journal      = {World Wide Web {(WWW)}},
  volume       = {29},
  number       = {2},
  pages        = {20},
  year         = {2026},
}

@inproceedings{tntood,
  author       = {Danny Wang and
                  Ruihong Qiu and
                  Guangdong Bai and
                  Zi Huang},
  title        = {Text Meets Topology: Rethinking Out-of-distribution Detection in Text-Rich Networks},
  booktitle    = {{EMNLP}},
  year         = {2025},
}

@article{cassette,
author = {Tang, Yanran and Qiu, Ruihong and Yin, Hongzhi and Li, Xue and Huang, Zi},
title = {Cassette: Case-to-Case Structural Distillation for Efficient Legal Case Retrieval},
year = {2026},
journal = {TOIS},
}

@inproceedings{caselink,
  author       = {Yanran Tang and
                  Ruihong Qiu and
                  Hongzhi Yin and
                  Xue Li and
                  Zi Huang},
  title        = {{CaseLink:} Inductive Graph Learning for Legal Case Retrieval},
  booktitle    = {{SIGIR}},
  year         = {2024},
}

@inproceedings{casegnn,
  author       = {Yanran Tang and
                  Ruihong Qiu and
                  Yilun Liu and
                  Xue Li and
                  Zi Huang},
  title        = {{CaseGNN:} Graph Neural Networks for Legal Case Retrieval with Text-Attributed Graphs},
  booktitle    = {{ECIR}},
  year         = {2024},
}

@article{puma,
  author       = {Yilun Liu and
                  Ruihong Qiu and
                  Yanran Tang and
                  Hongzhi Yin and
                  Zi Huang},
  title        = {{PUMA:} Efficient Continual Graph Learning for Node Classification With Graph Condensation},
  journal      = {{TKDE}},
  year         = {2025},
}

@inproceedings{host,
  author       = {Yan Jiang and
                  Ruihong Qiu and
                  Zi Huang},
  title        = {Does Homophily Help in Robust Test-time Node Classification?},
  booktitle    = {WSDM},
  year         = {2026},
}

@article{gfmate,
  author       = {Yan Jiang and
                  Ruihong Qiu and
                  Zi Huang},
  title        = {GFMate: Empowering Graph Foundation Models with Test-time Prompt Tuning},
  journal      = {CoRR},
  volume       = {abs/2605.14809},
  year         = {2026},
}

@article{tide,
  author       = {Danny Wang and
                  Ruihong Qiu and
                  Zi Huang},
  title        = {What Information Matters? Graph Out-of-Distribution Detection via
                  Tri-Component Information Decomposition},
  journal      = {CoRR},
  volume       = {abs/2605.13032},
  year         = {2026},
}

@inproceedings{gcondenser,
  author       = {Yilun Liu and
                  Ruihong Qiu and
                  Zi Huang},
  title        = {{GCondenser:} Benchmarking Graph Condensation},
  booktitle    = {{CIKM}},
  year         = {2025},
}

@inproceedings{cat,
  author       = {Yilun Liu and
                  Ruihong Qiu and
                  Zi Huang},
  title        = {{CaT:} Balanced Continual Graph Learning with Graph Condensation},
  booktitle    = {{ICDM}},
  year         = {2023},
}

\clearpage

\appendix

\section{Detailed Related Work}
\label{app:related}

\subsection{Reasoning on Text-Attributed Graphs}
\label{sec:related-llm-tag}

Existing LLM-based TAG methods generally assume a fixed neighbourhood context constructed before generation. Early approaches use LLMs as text encoders for nodes and labels, followed by graph-aware aggregation over the resulting embeddings~\cite{li2024zerog,he2024tape,liu2024oneforall,wang2025llmbp}. Subsequent methods integrate graph structure directly into the language model input space through soft graph embeddings, graph-language token interleaving, or graph-aware adapters~\cite{tang2024graphgpt,chen2024llaga,kong2025gofa}. A complementary line probes whether LLMs can reason over graph structure when graphs are serialised into natural language~\cite{wang2023nlgraph}, and shared benchmarks have emerged to evaluate graph-LLM methods across diverse tasks~\cite{li2024glbench,wu2025llmnodebed}. More recent work applies post-training to elicit explicit graph reasoning behaviours from LLMs. GraphWiz~\cite{chen2024graphwiz} instruction-tunes chain-of-thought~\cite{wei2022cot} reasoning traces for graph problems, while NPG-Muse~\cite{wang2025npgmuse} distils reasoning trajectories from larger models.

On TAG reasoning specifically, recent reinforcement learning approaches still rely on static neighbour selection. Graph-R1~\cite{wu2025graphr1} compresses local subgraphs into textual summaries before GRPO fine-tuning, while TRN-R1-Zero~\cite{liu2026trnr1zero} performs reasoning over randomly sampled neighbourhood subgraphs using a neighbour-aware reward objective. Across these methods, the neighbourhood context is selected before reasoning begins through heuristic rules such as full-subgraph concatenation, summarisation, or random sampling, so neighbour acquisition itself is never treated as a learnable decision process.

\subsection{Agentic Search}
\label{sec:related-interactive}

Recent agentic retrieval systems interleave reasoning with external information acquisition by allowing language models to issue search queries during generation, e.g., Self-RAG~\cite{asai2024selfrag}, Search-o1~\cite{li2025searcho1}, Search-r1~\cite{jin2025searchr1}, R1-Searcher~\cite{song2025r1searcher} and ReSearch~\cite{chen2025research}. These methods typically retrieve documents through external semantic retrievers, while the retrieval LLM is supervised only through final outcome rewards. The action space and environment are thus an unstructured document corpus resolved by similarity, rather than a graph-structured neighbourhood. This structural difference is load-bearing: a degree-preserving rewire that keeps every node's text and degree intact but destroys topology collapses walk accuracy to the preview+direct baseline (\S\ref{sec:walk-effectiveness}), so the gain is not recoverable by flat retrieval over the same node texts.

\subsection{Credit Assignment and On-Policy Self-Distillation}
\label{sec:related-opsd}

Sequential neighbour exploration introduces a sparse credit-assignment problem: a graph-walk trajectory may contain multiple neighbour-selection actions while receiving supervision only from a final task-level reward. Process reward models (PRMs)~\cite{lightman2024prm800k,wang2024mathshepherd} partially address sparse supervision in LLM reasoning by assigning step-level rewards to intermediate reasoning steps. However, PRMs require labelled intermediate trajectories or reference solutions, which are unavailable for graph exploration tasks. Recent work on on-policy self-distillation (OPSD) instead provides dense optimisation signals by comparing the LLM against a more-informed teacher LLM at the token level~\cite{shenfeld2026sdft,hubotter2026sdpo,zhao2026opsd}. Existing OPSD methods construct the teacher using privileged outcome information, such as gold answers~\cite{hubotter2026sdpo,zhao2026opsd}, reference solutions, or externally supplied hints~\cite{wang2026openclaw}. Consequently, the resulting supervision measures agreement with a known-correct continuation. CNY extends OPSD to graph exploration settings where no labelled intermediate supervision exists. Our destination-conditioned OPSD conditions the teacher on the revealed destination reached by the LLM itself, namely the neighbour retrieved by a graph-walk action.

\section{OPSD Credit Derivation}
\label{app:revkl-deriv}

This appendix derives the walk-action credit $\delta_t$ of Eq.~\eqref{eq:opsd-credit} as the gradient of a per-token reverse-KL self-distillation objective. Let $t$ index token positions in $\tau$, with $y_t$ the token at position $t$ and $s_{<t}$ the autoregressive prefix (the initial prompt plus all earlier tokens). The student and teacher next-token distributions at position $t$ are $\pi_\theta(\cdot \mid s_{<t})$ and $\pi_\theta(\cdot \mid s_{<t} \,\Vert\, \bar o_t)$, where $\bar o_t$ is the recap attached to the walk action containing position $t$ (undefined when no such action exists) and $\Vert$ denotes concatenation of the recap into $s_{<t}$ immediately before the action span; $\pi^{\mathrm{stu}}_t(y)$ and $\pi^{\mathrm{tea}}_t(y)$ denote the probability each assigns to a token $y$. At each walk-action token, OPSD distils the student towards its own destination-conditioned teacher by minimising the per-token reverse KL divergence
\begin{equation}
\begin{split}
\mathcal{D}^{\mathrm{rev}}_t \;\triangleq\;& \mathrm{KL}\!\big(\pi_\theta(\cdot \mid s_{<t}) \,\big\Vert\, \pi_\theta(\cdot \mid s_{<t} \,\Vert\, \bar o_t)\big) \\
\;=\;& \mathbb{E}_{y \sim \pi^{\mathrm{stu}}_t}\!\big[\log \pi^{\mathrm{stu}}_t(y) - \log \pi^{\mathrm{tea}}_t(y)\big],
\end{split}
\label{eq:revkl}
\end{equation}
the standard on-policy self-distillation objective: the student samples the token on policy and is pulled towards the recap-informed teacher, which is held fixed (stop-gradient)~\cite{shenfeld2026sdft,hubotter2026sdpo,zhao2026opsd}. Because the teacher is detached, the gradient is a REINFORCE-style gradient whose per-token weight is the teacher-to-student log-ratio,
\begin{equation}
\begin{split}
\nabla_\theta \mathcal{D}^{\mathrm{rev}}_t \;=\;& -\,\mathbb{E}_{y \sim \pi^{\mathrm{stu}}_t}\!\big[\,\delta_t(y)\,\nabla_\theta \log \pi^{\mathrm{stu}}_t(y)\,\big], \\
\delta_t(y) \;\triangleq\;& \log \pi^{\mathrm{tea}}_t(y) - \log \pi^{\mathrm{stu}}_t(y),
\end{split}
\label{eq:revkl-grad}
\end{equation}
so minimising the reverse KL is identical to using $\delta_t$ as a per-token advantage~\cite{shenfeld2026sdft,hubotter2026sdpo}: crediting the action by the log-ratio and distilling towards the teacher are the same update. Evaluated at the realised token $y_t$, this gives the single-sample credit $\delta_t = \log \pi^{\mathrm{tea}}_t - \log \pi^{\mathrm{stu}}_t = -\mathrm{KL}^{\mathrm{rev}}_t$ of Eq.~\eqref{eq:opsd-credit}, with $\mathrm{KL}^{\mathrm{rev}}_t = \log \pi^{\mathrm{stu}}_t - \log \pi^{\mathrm{tea}}_t$ the per-token reverse log-ratio. As a single realised-token term it is a biased estimator of the reverse-KL gradient~\cite{shenfeld2026sdft}, used directly as the dense credit.

\section{Algorithms}
\label{app:algorithms}

Algorithm~\ref{alg:cny-rollout} gives one CNY rollout (\S\ref{sec:rollout}); Algorithm~\ref{alg:cny-train} gives one training step (\S\ref{sec:grpo}), pricing the GRPO advantage with the OPSD credit of Eq.~\eqref{eq:opsd-adv}.

\begin{algorithm}[t]
\caption{CNY rollout}
\label{alg:cny-rollout}
\small
\begin{algorithmic}[1]
\Require target node $v_0$, LLM $\pi_\theta$, max hops $T_{\max}$
\State $q_0 \gets \textsc{BuildPrompt}(v_0, \mathcal{N}(v_0), \mathcal{Y})$
\State $h_0 \gets \varepsilon$, $\mathcal{F}_0 \gets \mathcal{N}(v_0)$
\For{$k = 1 \dots T_{\max}$}
  \State $a_k \sim \pi_\theta(\cdot \mid q_0, h_{k-1})$
  \If{$a_k = \texttt{<walk>}X\texttt{</walk>}$ and $X \in \mathcal{F}_{k-1}$}
    \State $o_k \gets \texttt{<information>}\,x_X\,\texttt{</information>}$
    \State $h_k \gets h_{k-1} \Vert a_k \Vert o_k$
    \State $\mathcal{F}_k \gets \textsc{UpdateFrontier}(\mathcal{F}_{k-1}, X)$ \Comment{Eq.~\ref{eq:frontier-update}}
  \ElsIf{$a_k = \texttt{<answer>}c\texttt{</answer>}$}
    \State \Return $c$, $\tau = (q_0, h_{k-1} \Vert a_k)$
  \Else
    \State $h_k \gets h_{k-1} \Vert a_k$ \Comment{malformed; loop continues}
  \EndIf
\EndFor
\State \Return $\bot$
\end{algorithmic}
\end{algorithm}

\begin{algorithm}[t]
\caption{CNY training step}
\label{alg:cny-train}
\small
\begin{algorithmic}[1]
\Require batch $\mathcal{B}$, group size $N$, OPSD coefficient $\beta$

\For{$v_0 \in \mathcal{B}$}
    \State sample $\tau_1,\ldots,\tau_N$ in parallel via Alg.~\ref{alg:cny-rollout}
    \State $r_i \gets R(\tau_i)$ \Comment{Eq.~\ref{eq:reward-tier}}
    \State $\{\hat A_{\tau_i}\}_{i=1}^{N}
    \gets
    \textsc{GroupNormalise}(\{r_i\}_{i=1}^{N})$
    \Comment{GRPO}
    
    \For{$i = 1 \dots N$, each walk action $\texttt{<walk>}X\texttt{</walk>}$ with observation $o$ in $\tau_i$}
        \State $\bar o \gets \textsc{Recap}(o)$
        \State $\tau_i^{\mathrm{tea}}
        \gets
        \textsc{InsertDestinationRecap}(\tau_i,\bar o)$
        
        \State compute
        $\log \pi_t^{\mathrm{stu}}$
        and
        $\log \pi_t^{\mathrm{tea}}$
        on the node-ID tokens of $X$
        
        \State
        $\delta_t
        \gets
        \log \pi_t^{\mathrm{tea}}
        -
        \log \pi_t^{\mathrm{stu}}$
        \Comment{Eq.~\ref{eq:opsd-credit}}
        
        \State
        $\tilde A_t
        \gets
        \hat A_{\tau_i}
        +
        \beta\,\delta_t$
        \Comment{Eq.~\ref{eq:opsd-adv}}
    \EndFor
    
    \State
    $\tilde A_t
    \gets
    \hat A_{\tau_i}$
    for all remaining tokens in $\tau_i$
\EndFor

\State
$\theta
\gets
\theta
-
\eta
\nabla
\mathcal L_{\mathrm{PPO\mbox{-}clip}}
(\{\tilde A_t\};
\epsilon_{\mathrm{lo}},
\epsilon_{\mathrm{hi}})$

\end{algorithmic}
\end{algorithm}

\section{Datasets and Prompt Templates}
\label{app:datasets}

This appendix documents the datasets and the prompt templates that turn each graph instance into a rollout. All stochastic data construction uses a fixed random state, so rebuilding from source reproduces the exact splits, neighbourhoods and rollouts used in every experiment.

\subsection{Baselines and Evaluation Details}
\label{app:setup-details}

The evaluation suite reuses Graph-R1's Table~1 datasets~\cite{wu2025graphr1}, omitting the molecular regression subset and link prediction. Expla-Graph is the sole graph-level evaluation, and its per-instance explanation graph supplies a walkable concept neighbourhood evaluated under the same walk template as node classification (App.~\ref{app:expla-graph}).

Within the three baseline families of \S\ref{sec:setup}, the general-purpose LLMs are Llama2-7B~\cite{touvron2023llama2} and Mistral-7B~\cite{jiang2023mistral}; the graph foundation models are OFA~\cite{liu2024oneforall}, UniGraph~\cite{he2025unigraph}, LLaGA~\cite{chen2024llaga} and GOFA~\cite{kong2025gofa} (in its GOFA-T and GOFA-F settings); numbers for both families are taken from the Graph-R1 benchmark. Graph-R1's reported results summarise node texts with DeepSeek-v3 before reasoning; following TRN-R1-Zero, we re-evaluate Graph-R1 from its released checkpoint on raw node text instead.

\subsection{Graph Statistics}
\label{app:dataset-stats}

Table~\ref{tab:dataset-stats} reports the raw graph for each dataset. Edge counts are reported directly from the stored edge index; for undirected citation and co-purchase graphs each undirected edge is counted twice. For the WN18RR edge split the reported sizes count edges, the classification unit, rather than nodes.

\begin{table*}[t]
\centering
\setlength{\tabcolsep}{3pt}
\begin{tabular*}{\textwidth}{@{\extracolsep{\fill}}lrrrrrr@{}}
\toprule
Dataset & Nodes & Edges & Classes & Train & Val & Test \\
\midrule
CiteSeer        &   3{,}186 &      8{,}450 &  6 &   1{,}911 &     637 &     638 \\
PubMed          &  19{,}717 &     88{,}648 &  3 &  11{,}830 &  3{,}944 &  3{,}943 \\
Photo           &  48{,}362 &    873{,}782 & 12 &  29{,}016 &  9{,}674 &  9{,}672 \\
Computer        &  87{,}229 &  1{,}256{,}548 & 10 &  52{,}336 & 17{,}446 & 17{,}447 \\
History         &  41{,}551 &    503{,}180 & 12 &  24{,}930 &  8{,}310 &  8{,}311 \\
Sportsfit       & 173{,}055 &  3{,}020{,}134 & 13 & 103{,}834 & 34{,}610 & 34{,}611 \\
Instagram       &  11{,}339 &    144{,}010 &  2 &   6{,}803 &  2{,}268 &  2{,}268 \\
WN18RR          &  40{,}943 &     93{,}003 & 11 &  86{,}835 &  3{,}034 &  3{,}134 \\
\midrule
Cora            &   2{,}708 &     10{,}556 &  7 &   1{,}626 &     542 &     540 \\
WikiCS          &  11{,}701 &    431{,}206 & 10 &   7{,}020 &  2{,}341 &  2{,}340 \\
Products        &  54{,}025 &     74{,}420 & 44 &  14{,}708 &  1{,}572 & 37{,}745 \\
FB15K237        &  14{,}541 &    310{,}116 & 237 & 272{,}115 & 17{,}535 & 20{,}466 \\
Expla-Graph     &  14{,}294 &     23{,}490 &   2 &   1{,}659 &      553 &      554 \\
\bottomrule
\end{tabular*}

\caption{\textbf{Dataset statistics.}}
\label{tab:dataset-stats}
\end{table*}

The top block lists training datasets, the bottom block evaluation datasets that never enter the training corpus. The Products label vocabulary lists 47 names, of which only 44 occur as ground-truth classes.

\subsection{Node-Classification Prompt}
\label{app:prompt-template}

Every node-classification prompt follows a single multi-step template. In place of an explicit output-format specification, the template provides two illustrative example turns and relies on the strict format reward (\S\ref{sec:rollout}) to enforce well-formed spans. Figure~\ref{fig:prompt-compact} shows the template skeleton with placeholder fields; Figure~\ref{fig:prompt-v2} renders one Cora test instance verbatim, with example node IDs and label indices substituted per instance.

\begin{figure}[t]
\centering
\input{figures/prompt_v2_compact}
\caption{\textbf{CNY rollout prompt template.} Curly-brace fields are substituted per node.}
\label{fig:prompt-compact}
\end{figure}

\begin{figure}[t]
\centering
\input{figures/recap_prompt}
\caption{\textbf{Destination recap prompt used by OPSD} (\S\ref{sec:opsd}). The recap is produced by the LLM itself and constrained to a single topic-only sentence of at most 50 tokens, with node IDs, walk recommendations and any guess at the target's class explicitly forbidden, so the recap supplies destination context without leaking the gold label or a navigation hint.}
\label{fig:recap-prompt}
\end{figure}
\begin{figure}[t]
\centering
\input{figures/prompt_v2}
\caption{\textbf{CNY node-classification prompt.} A Cora test instance; node and label IDs are substituted per instance.}
\label{fig:prompt-v2}
\end{figure}

\subsection{Edge-Classification Prompt}
\label{app:prompt-edge}

Edge classification (FB15K237, WN18RR) reuses the same multi-step walk template: the target slot carries two anchors (head and tail entity) rather than one, and the initial frontier is the union of their 1-hop neighbourhoods, so a walk may inspect either side. The label slot is the relation vocabulary, shortlisted to $K{=}10$ per edge to match the n-way evaluation column reported in Table~\ref{tab:main} (gold relation plus $9$ sampled distractors, drawn deterministically from the FB15K237 relation vocabulary). The walk budget is non-trivial ($\texttt{max\_hops}{=}5$): naming a Freebase relation typically requires reading at least one neighbour on each side. Figure~\ref{fig:prompt-fb15k237} shows one FB15K237 test instance verbatim; entity IDs and relation indices are substituted per edge.

\begin{figure}[t]
\centering
\input{figures/prompt_fb15k237}
\caption{\textbf{CNY edge-classification prompt.} An FB15K237 test instance. Two anchors (head, tail) replace the single target, and the label slot is a $10$-way shortlist over the FB15K237 relation vocabulary; entity and relation IDs are substituted per edge.}
\label{fig:prompt-fb15k237}
\end{figure}

\subsection{Expla-Graph Stance-Walk Construction}
\label{app:expla-graph}

Expla-Graph (\S\ref{sec:main-results}) is mapped onto the same multi-step walk template used for node classification, so the evaluation prompt stays in-distribution: the (belief, argument) pair occupies the target slot, the two stances ($0$ counter, $1$ support) occupy the category slot, and the explanation graph supplies a walkable concept neighbourhood. Every instance's concept graph is packed into one disjoint global graph, so a \texttt{<walk>} can never leave its own instance.

\paragraph{Walkable neighbourhood.} The initial frontier is a connected seed subset of the concept graph, a seed concept together with its one-hop concepts, shown as name-only previews; the remaining concepts stay hidden until walked. Each concept's relations live in its node text and surface through the existing \texttt{<information>} channel after a walk (``\texttt{Node X: <name>. Connections: ...}''), so no edge text or bespoke block is introduced.

\paragraph{Stance decision rule.} The task-instruction slot carries a stance rule, ``first determine what the belief claims, then judge whether the argument agrees with that claim (support) or disagrees with it (counter)''. The dominant error without it is judging whether the argument is positive about the topic rather than whether it agrees with the belief, which inverts the label whenever the belief is itself a contrarian claim. The rule occupies the generic task-instruction slot and the label names stay plain (\texttt{counter}, \texttt{support}), so scoring is untouched.

\paragraph{Leakage control.} The gold explanation-graph string is never placed in the prompt, as it trivially discloses the stance; only the belief, the argument, the concept names (previews) and the relations revealed on a walk are surfaced. Figure~\ref{fig:prompt-expla-graph} shows one evaluation instance verbatim; the concept IDs and stance indices are substituted per instance.

\begin{figure}[t]
\centering
\input{figures/prompt_expla_graph}
\caption{\textbf{CNY graph-reasoning prompt.} An Expla-Graph test instance (gold stance: support). The (belief, argument) pair is the target and the explanation-graph concepts are the walkable neighbourhood; concept IDs and stance indices are substituted per instance.}
\label{fig:prompt-expla-graph}
\end{figure}

\subsection{Destination Recap Prompt}
\label{app:recap-prompt}

The OPSD credit of \S\ref{sec:opsd} re-scores each walk action under a recap of the destination it reached. Figure~\ref{fig:recap-prompt} gives the prompt that produces this recap. The recap is generated by the LLM itself and constrained to a single topic-only sentence: node IDs, walk recommendations and any guess at the target's class are explicitly forbidden, so the recap supplies destination context without leaking the gold label or a navigation hint.

\section{Implementation}
\label{app:implementation}

Table~\ref{tab:hyperparams} lists the optimisation and rollout hyperparameters of the CNY-14B reference run; Table~\ref{tab:hardware} lists the compute allocation per backbone. All values are read from the released training configuration.

\begin{table}[t]
\centering
\small
\setlength{\tabcolsep}{4pt}
\resizebox{\columnwidth}{!}{%
\begin{tabular}{ll}
\toprule
Setting & Value \\
\midrule
Backbone & Qwen2.5-14B-Instruct \\
GRPO group size $N$ (rollouts per prompt) & 8 \\
Prompt batch size & 32 \\
Global batch size & 256 \\
Learning rate & $1\times10^{-6}$ (constant) \\
Optimiser & Adam ($\beta_1{=}0.9$, $\beta_2{=}0.98$) \\
Weight decay & $0.01$ \\
PPO clip $\epsilon_{\text{low}}/\epsilon_{\text{high}}$ & $0.2 / 0.28$ \\
OPSD coefficient $\beta$ & $0.03$ \\
Max hops $T_{\max}$ & 5 \\
Initial frontier cap (training) & 3 \\
Recap length $L_\varphi$ & $\le 50$ tokens \\
Max response length & 2{,}048 tokens \\
Context length & 32{,}768 tokens \\
Rollout steps & 1{,}000 \\
\bottomrule
\end{tabular}%
}

\caption{\textbf{Hyperparameters of the CNY-14B reference run} (Qwen2.5-14B-Instruct, GRPO with OPSD), as read from the released training configuration.}
\label{tab:hyperparams}
\end{table}
\begin{table}[t]
\centering
\small
\setlength{\tabcolsep}{4pt}
\begin{tabular}{lll}
\toprule
Backbone & GPUs & Parallelism \\
\midrule
Qwen2.5-14B, Qwen3-14B & $4\times$ H100 & TP$=4$ \\
Qwen2.5-7B & $2\times$ H100 & TP$=2$ \\
Llama-3.2-3B, Qwen3-4B & $1\times$ H100 / L40 & TP$=1$ \\
\bottomrule
\end{tabular}

\caption{\textbf{Compute allocation per backbone.} The smallest backbones also fit a single L40; the headline CNY-14B run uses four H100 80GB GPUs under Megatron tensor parallelism.}
\label{tab:hardware}
\end{table}

On the headline configuration of Table~\ref{tab:hardware} ($4\times$H100 80GB, TP$=4$), the 14B run trains at a median of ${\sim}300$\,s per step (${\sim}280$\,s on a typical non-evaluation step; evaluation runs every tenth step), under a $1{,}000$-step wall limit beyond which training halts; the schedule fits within each H100's 80\,GB under TP$=4$.

\section{Additional Analyses}
\label{app:additional}

All measurements below use the frozen CNY-14B unless a training run is stated.

\subsection{Preview Construction}
\label{app:preview-spec}

Previews are rendered as \texttt{[\{id\}]: \{prefix\} ...}, with per-dataset lengths fixed in the released configuration; Expla-Graph is the sole exception, where a preview is the concept name (App.~\ref{app:expla-graph}). Varying the budget at inference only, WikiCS accuracy is $77.1$ at $5$ tokens (title-only, as the page title opens each node's text), $77.9$ at $10$, $76.8$ at the published $30$ and $77.5$ at $60$.

\subsection{Matched Preview Exposure}
\label{app:preview-exposure}

{\color{black}
The walk arm also sees lightweight previews of the whole neighbourhood, which could itself drive the gain. We therefore let the model answer directly while shown previews of every neighbour but forbidden to walk, matched to each dataset's preview length and frontier size. Exposing all previews improves the direct baseline only modestly (Cora $65.0\!\to\!66.3$, WikiCS $71.0\!\to\!73.4$, Products $83.0\!\to\!83.5$), and walking still adds a substantial margin over this stronger control (Cora $66.3\!\to\!74.1$, WikiCS $73.4\!\to\!76.6$, Products $83.5\!\to\!87.6$), placing the gain in reading the selected neighbour, not in preview exposure.
}

\subsection{Failure Modes of Walking}
\label{app:failure-modes}

A flip is a node that the target text alone classifies correctly but walking then misclassifies. Flips are pooled over three walk-evaluation seeds and assigned to the three modes of \S\ref{sec:case-study} by manual inspection. Within the first mode they concentrate on adjacent class pairs, most often \texttt{Computer Architecture} to \texttt{Operating Systems}, and the second mode collects multi-facet targets whose destination text foregrounds the wrong facet.

\end{document}